\documentclass[a4paper,fleqn]{cas-dc}

\usepackage[numbers]{natbib}
\usepackage{graphicx}
\usepackage{CJKutf8}
\usepackage{capt-of} 
\usepackage{tcolorbox}
\usepackage{booktabs}    
\usepackage{tabularx}    
\def\tsc#1{\csdef{#1}{\textsc{\lowercase{#1}}\xspace}}
\tsc{WGM}
\tsc{QE}
\tsc{EP}
\tsc{PMS}
\tsc{BEC}
\tsc{DE}

\begin{document}
\let\WriteBookmarks\relax
\def\floatpagepagefraction{1}
\def\textpagefraction{.001}

\shorttitle{}


\title [mode = title]{From Business Events to Auditable Decisions: Ontology-Governed Graph Simulation for Enterprise AI}

\author[1,2]{Hongyin Zhu}[orcid=0000-0001-5786-7594]\cormark[1]
\ead{zhuhongyin@yonyou.com}
\author[2]{Jinming Liang}
\author[2]{Mengjun Hou}
\author[1,2]{Ruifan Tang}
\author[1,2]{Xianbin Zhu}
\author[2]{Jingyuan Yang}
\author[2]{Yuanman Mao}
\author[2]{Feng Wu}
\affiliation[1]{organization={Yonyou AI Lab}}

\affiliation[2]{organization={Yonyou Network Technology Co., Ltd.},
    }

\begin{abstract}
Existing LLM-based agent systems share a common architectural failure: they answer
from the unrestricted knowledge space without first simulating how active business
scenarios reshape that space for the event at hand---producing decisions that are
fluent but ungrounded and carrying no audit trail.
We present LOM-action, which equips enterprise AI with \emph{event-driven ontology
simulation}: business events trigger scenario conditions encoded in the enterprise
ontology~(EO), which drive deterministic graph mutations in an isolated sandbox,
evolving a working copy of the subgraph into the scenario-valid simulation
graph $G_{\text{sim}}$; all decisions are derived exclusively from this evolved graph.
The core pipeline is \emph{event $\to$ simulation $\to$ decision}, realized through
a dual-mode architecture---\emph{skill mode} and \emph{reasoning mode}. Every decision produces a fully traceable audit log.
LOM-action achieves 93.82\% accuracy and 98.74\% tool-chain F1 against frontier
baselines Doubao-1.8 and DeepSeek-V3.2, which reach only 24--36\% F1 despite 80\%
accuracy---exposing the \emph{illusive accuracy} phenomenon. The four-fold F1 advantage
confirms that ontology-governed, event-driven simulation, not model scale, is the
architectural prerequisite for trustworthy enterprise decision intelligence.
\end{abstract}


\maketitle

\section{Introduction}

Enterprise AI cannot be built on general-purpose LLMs alone.
Enterprise decisions are not made on the static ontology---they are made on a
scenario-evolved version of it, shaped by the active business conditions of the event
at hand: the carrier contracts in force, the spending policy currently active, the
organizational scope of the requesting user.
General-purpose LLMs have no mechanism to perform this evolution: they answer from the
unrestricted knowledge space, producing responses that are fluent but never derived from
the graph the business scenario actually defines.

LOM~\cite{zhu2026unifying,zhang2026construct} established the ontological foundation
for enterprise AI.
This paper extends LOM with a new capability that production deployment demands:
a \emph{sandbox simulation engine} that evolves a working copy of the enterprise
ontology under active business scenario conditions, and derives every decision
exclusively from the resulting simulation-valid graph.

LOM-action adds the next piece of the capability puzzle: \emph{knowing how to delegate}.
Where LOM grounds what the model knows and how it reasons, LOM-action governs what the
model can invoke---a registry of external resources including frontier models, specialized
tool endpoints, and domain skill nodes, each accessed through a skill ontology node
carrying EO authorization contracts, making every delegation an auditable ontological
act rather than a black-box API call.
A frontier LLM invoked as a registered skill node operates on a precisely bounded,
EO-authorized input; its output is intercepted by LOM-as-Judge, which re-grounds the
result against EO authority before writing it to the session ontology.
The capability ceiling therefore rises continuously as the registered skill set expands
and frontier model quality improves, without retraining.

Each business event---a structured data payload from an enterprise system that carries
sufficient semantic content to activate EO-encoded scenario conditions---poses a
specific question: not ``what does the static graph say?'' but ``what does the graph say
after this event's scenario conditions have reshaped it?''
Event-driven AI answers by first evolving the ontology under those conditions,
then deriving decisions exclusively from the evolved state.

Business scenarios are properties of the ontological context, not features of event
text, so injecting them as prompt instructions fails: the model treats them as
soft preferences and may still operate on the unrestricted graph.
The correct scope for any tool call is not the static graph but the simulation-valid
$G_{\text{sim}}$ that survives the organization's active scenario conditions; a system
that bypasses simulation answers a different question than the one the enterprise event
posed, with no audit basis.
LOM-action addresses both gaps with a strictly ordered three-phase pipeline---scenario
parsing, sandbox simulation, decision derivation---backed by a persistent, isolated
graph sandbox: when a business event arrives, a working copy of the enterprise ontology is instantiated under a unique \texttt{graph\_id}; Phase~2 mutates this copy
without touching the authoritative EO graph; Phase~3 executes decisions against the
evolved state, with every mutation logged for full auditor replay.

This paper makes three contributions.
(1) The \emph{scenario simulation} innovation: EO-authorized constraint predicates drive
deterministic sandbox graph mutations before any decision is derived, confirmed
empirically to close the simulation gap that frontier LLMs systematically leave open.
(2) The \emph{decision derivation} innovation: the event $\to$ simulation $\to$ decision
pipeline via a dual-mode architecture---skill mode for registered skill calls; reasoning
mode for novel computations---with every decision producing a fully traceable,
replayable decision trace.
(3) The \emph{simulation-first principle} and the illusive accuracy index
$\text{IA}(M) = \text{Acc}(M) - \text{F1\_chain}(M)$, validated across 11 tasks.

The remainder of this paper is organized as follows.
Section~2 reviews related work.
Section~3 presents the approach across Sections~3.1--3.7.
Section~4 reports experiments, results, and limitations.
Section~5 concludes.
Appendix~A covers the ontology harness engineering and broader LOM global architecture.

\section{Related Work}
\subsection{Tool-Augmented LLM Agents}
The modern tool-use paradigm traces from ReAct~\cite{yao2022react}, which interleaves
reasoning traces with action execution, through
Toolformer~\cite{schick2023toolformer}, which introduces self-supervised tool
annotation, to OpenAI function calling~\cite{hurst2024gpt}, which standardized JSON
Schema-based API invocation as a first-class model interface.
Subsequent work has pushed this paradigm in two directions: scale and specialization.
On scale, frontier systems---GPT-4o~\cite{hurst2024gpt}, Claude with extended
thinking~\cite{anthropic2024claude3oct}, Gemini function
calling~\cite{team2023gemini}---have dramatically raised the zero-shot ceiling for tool
invocation quality.
On specialization, fine-tuning methods such as ToolACE~\cite{liu2024toolace},
Hammer~\cite{lin2024hammer}, and xLAM~\cite{zhang2025xlam} demonstrate that
domain-curated corpora can substantially close the gap with frontier models on standard
function-calling benchmarks.
Evaluation has matured accordingly: BFCL~\cite{patil2025berkeley} provides a rigorous
leaderboard covering parallel, nested, and multi-turn invocations;
$\tau$-bench~\cite{yao2024tau} evaluates agentic task completion across extended
dialogues; WorFBench~\cite{qiao2024benchmarking} targets multi-step workflow planning.

Despite this progress, all these systems share one architectural assumption: tool
selection is a language model inference decision made from the full input context, with
scenario conditions provided (if at all) as natural language instructions in the prompt.
No system introduces a mandatory sandbox simulation step that evolves the knowledge
graph before any tool is invoked.
LOM-action challenges this at the root: tool selection is a consequence of
sandbox-simulated graph evolution, not a generative act over the unrestricted space.
Existing benchmarks also share a measurement gap---they evaluate tool-call accuracy
without measuring whether decisions were derived from a correctly evolved graph.
Our tool-chain F1 and Illusive Accuracy index fill this gap by distinguishing decisions
produced through scenario-driven simulation from those produced by operating on the
unrestricted graph, exposing a failure mode invisible to standard accuracy metrics.

\subsection{Knowledge Graph Reasoning and Enterprise Semantic Systems}
The integration of LLMs with ontologys has advanced rapidly along the retrieval
axis.
KG-augmented LLMs embed graph information---including domain-specific heterogeneous
knowledge unified into structured representations~\cite{zhu2023pre}---in prompts or
fine-tuning data to improve factual grounding.
The underlying ontologys themselves depend on structured relation extraction
pipelines~\cite{zhu2022switchnet} that adaptively identify typed entity relationships
in text, a foundational step that KG-augmented LLM approaches typically assume as a
precondition.
GraphRAG~\cite{edge2024local} introduces community-aware hierarchical graph
summarization as a retrieval layer for document corpora.
GNN-RAG~\cite{mavromatis2024gnn} combines graph neural network traversal with
retrieval-augmented generation for multi-hop question answering.
Think-on-Graph~2.0~\cite{ma2024think} iteratively beam-searches ontology paths
to guide LLM reasoning.
Complementary to these retrieval-focused methods, graph representation learning has
advanced node-level understanding through semantic-structural attention-enhanced graph
convolutional networks~\cite{zhu2024node} and pre-trained graph autoencoders
incorporating hierarchical topology knowledge~\cite{zhu2025retracted}.

Across all these works, the ontology is a \emph{static retrieval substrate}---a
source of evidence that the LLM draws on but does not modify.
Enterprise-specific systems follow a similar pattern: text-to-SQL systems for
BI~\cite{floratou2024nl2sql,gao2023text} treat database schemas as retrieval aids, and
DB-GPT~\cite{xue2023db} integrates LLMs with database engines for natural language
querying but positions the schema as context rather than authority.
None of these systems maintains a mutable simulation sandbox where the graph is evolved
to model business scenarios before decisions are derived; the ontology is
consulted, not simulated.
LOM-action differs fundamentally: the sandbox is a dynamic, mutable execution workspace
where scenario operations evolve a copy of the enterprise graph in an isolated
environment, subsequent reasoning operates on the evolved state, and the EO is not
merely consulted---it authorizes every mutation before any model inference proceeds.
The simulation-first principle---the best decision is the best decision among those
derivable from the simulation-valid graph---has no direct precedent in the enterprise
LLM literature.

\subsection{Long-Context Management}
The long-context LLM literature has approached the problem of extended conversations
primarily through capacity: Gemini~1.5 Pro supports up to 1M
tokens~\cite{team2024gemini}, GPT-4o handles 128K, and architectural innovations such
as Longformer~\cite{beltagy2020longformer} and Transformer-XL~\cite{dai2019transformer}
extend effective attention spans.
Complementary work on context
efficiency---SnapKV~\cite{li2024snapkv}, PyramidKV~\cite{cai2024pyramidkv},
LLMLingua~\cite{pan2024llmlingua}---compresses or prunes token sequences to improve
utilization within fixed windows.
These approaches treat the context management problem as one of capacity: fit more
history, or compress it more efficiently.

LOM-action reframes the problem as one of \emph{semantic precision}: the issue is not
that context windows are too short, but that accumulated raw text is semantically
undifferentiated---every prior turn competes for the same attention budget regardless
of relevance to the current event.
By replacing raw conversation history with typed session ontology (SO) subgraph deltas
keyed to EO entities and relations, LOM-action achieves session-positional
invariance---the same event type on the same ontological state produces the same simulation-grounded decision regardless of conversational position, because context
footprint is bounded by semantic scope rather than turn count.
This is a property that capacity scaling cannot provide.

\subsection{Deployed Enterprise AI Products}
A distinct competitive landscape consists of AI systems already deployed in enterprise
settings: Microsoft Copilot for Microsoft~365 (released 2023) integrates
LLM generation into productivity workflows via retrieval over document corpora and
graph-based organizational context from Microsoft~Graph; Salesforce Einstein~GPT
(2023) embeds LLMs into CRM event pipelines, enabling natural language interactions over
customer records and workflow triggers; ServiceNow AI (2024) integrates generative and
predictive models into IT service management workflows with event-driven ticket routing.
These systems demonstrate the commercial viability of event-triggered AI in enterprise
contexts and confirm the demand for decision automation at scale.

LOM-action's contribution is architecturally orthogonal to these deployments.
The systems above are primarily \emph{retrieval-augmented}: they fetch relevant
documents or records and provide them as context to a general-purpose LLM, which then
generates a response over the full retrieved context.
None introduces a mandatory ontology-governed simulation step that evolves the
enterprise ontology under active scenario conditions before any decision is
derived; none enforces EO-authorized constraint predicates as a structural gate on the
reasoning substrate; and none produces a replayable, operation-level decision trace
whose scope is provably bounded to the simulation decision graph $G_{\text{sim}}$.
Where those systems ask ``what does the retrieved context say?'', LOM-action asks
``what does the scenario-evolved graph say?''---a distinction that is invisible to
retrieval accuracy metrics but directly measurable through tool-chain F1.

\section{Approach}

\subsection{Scenario Simulation and Decision Derivation}

LOM-action is defined by two innovations that form a unified pipeline, not independent
contributions.

\textbf{Innovation~1 --- Scenario Simulation.}
Enterprise events do not operate on the unrestricted ontology.
They operate within a context of business scenarios: EO-authorized conditions on
entities and relations that define which portion of the graph is the valid reasoning
substrate for this event, by this user, under the currently active organizational
policies.
Let $G = (V, E)$ be the enterprise ontology and $\mathcal{R} = \{r_1, \ldots, r_k\}$
the active scenario condition set.
Scenario conditions fall into two classes, producing distinct simulation graphs that
are collectively denoted $G_{\text{sim}}$.
\emph{Constraint conditions} restrict the graph by removing nodes or edges that violate
an access policy or organizational rule; they produce the scenario-valid subgraph
$G_{\mathcal{R}}$:
\begin{align*}
G_{\mathcal{R}} &= (V_{\mathcal{R}},\ E_{\mathcal{R}}), \\
V_{\mathcal{R}} &= \{v \in V \mid \forall r_i \in \mathcal{R},\ r_i(v) = \top\}, \\
E_{\mathcal{R}} &= E[V_{\mathcal{R}}]
\end{align*}
\emph{Augmentation conditions} extend the graph by adding new nodes and edges (e.g.,
newly created organizational units) or reweighting existing edges (e.g., applying a
surcharge schedule); they produce the scenario-augmented graph $G_{\mathcal{A}}$:
\begin{align*}
G_{\mathcal{A}} &= (V \cup \Delta V,\ E[V \cup \Delta V] \cup \Delta E)
\end{align*}
where $\Delta V$ and $\Delta E$ are the EO-authorized node and edge deltas produced by
the scenario program.
In both cases $G_{\text{sim}} \in \{G_{\mathcal{R}}, G_{\mathcal{A}}\}$ is the
simulation object: the working copy of the enterprise graph evolved under active
scenario conditions before any decision is derived.
Tasks~9--10 in our benchmark instantiate the constraint case ($G_{\mathcal{R}}$);
Task~11 instantiates the augmentation case ($G_{\mathcal{A}}$).
For brevity, the remainder of this paper uses $G_{\mathcal{R}}$ to refer to the
simulation graph in contexts where the constraint case is primary; augmentation-specific
properties are noted explicitly where they differ.
Every condition in $\mathcal{R}$ must be EO-authorized---a reimbursement ceiling cannot
be an arbitrary value; an access scope cannot be inferred from statistical patterns.
LOM-action supports not only simple attribute predicates but also complex natural
language-described business scenarios whose logic involves multi-step conditional
computation.
Such scenarios are parsed in Phase~1 into ordered sequences of Phase~2 sandbox
operations that implement the scenario's conditional logic as a deterministic graph
computation, with every intermediate state written to the SO evidence chain and every
value traced to an EO-authorized constraint.

\textbf{Innovation~2 --- Decision Derivation.}
In existing agent systems, tool selection is the first-class decision, made from the full input context.
In LOM-action, by the time the decision step executes, the information space has already
been reduced to $G_{\mathcal{R}}$ through sandbox simulation.
The tool call is not a decision about what to retrieve---it is a computation on an
already-simulated, already-scenario-valid graph.
The pipeline is strictly sequential:
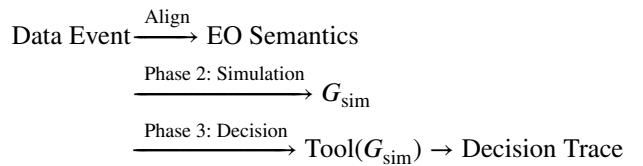
\begin{align*}
\text{Data Event} &\xrightarrow{\text{Align}} \text{EO Semantics} \\
&\xrightarrow{\text{Phase 2: Simulation}} G_{\text{sim}} \\
&\xrightarrow{\text{Phase 3: Decision}} \text{Tool}(G_{\text{sim}}) \to \text{Decision Trace}
\end{align*}
Skipping Phase~2 does not produce a faster answer---it produces an answer to the wrong
question.
The correct reasoning substrate for any decision is not the static graph: it is
$G_{\text{sim}}$ ($G_{\mathcal{R}}$ in the constraint case, $G_{\mathcal{A}}$ in the
augmentation case), the graph evolved by the organization's active scenario conditions.
A system that bypasses Phase~2 answers ``what does the static graph say?'' rather than
``what does the simulation-evolved graph say?''---and the resulting decision carries no
simulation trace and no audit basis.
Our experiments confirm this: F1 = 0.00 on basic traversal tasks for frontier baselines
despite near-perfect accuracy, and a 34\% accuracy gap on scenario tasks where
$G_{\mathcal{R}}$ differs substantively from~$G$.

\subsection{Enterprise AI vs.\ Consumer AI: The Simulation-First Principle}

\textbf{Prerequisite: the enterprise ontology.}
LOM-action's simulation pipeline assumes that an enterprise ontology (EO) exists and
is sufficiently populated to authorize the scenario conditions required by incoming
events.
This is a real prerequisite, not a trivial one.
Constructing and maintaining an enterprise ontology is a substantial undertaking that
precedes LOM-action deployment.
We addressed this issue in our prior work \cite{zhu2026unifying}.
The RAC (Reason $\to$ Align $\to$ Construct) evolutionary flywheel described in
Appendix~A.8 is our architectural path toward reducing the cold-start burden: by
capturing candidate ontology nodes from deployment interactions and routing them to a
governed review process, RAC enables the EO to grow incrementally from real
organizational use rather than requiring full pre-specification.
For the current instantiation, the 19-function graph API suite operates on synthetic
Neo4j ontologies whose nodes and scenario conditions are fully specified; the path to
real enterprise ontologies is through SKILLS-standard integration, which we identify as
the primary future work direction.

\textbf{The architectural inversion.}
In consumer AI the LLM holds full reasoning authority; everything else is scaffolding.
In enterprise AI this is inverted: the ontology is the authority and the LLM is one
component of the execution harness that channels ontological authority into natural
language interaction and decision derivation.
The full harness design is elaborated in Section~3.7; here we focus on the
optimization-theoretic implication of this inversion.

\textbf{Two optimization contracts.}
Consumer AI operates under a single-stage contract: maximize answer quality over the
full knowledge space.
There is no hard correctness boundary; approximate answers are acceptable.
Enterprise AI operates under a two-stage contract: (1) identify the simulation-valid
set $\mathcal{F}$---the set of decisions derivable from the scenario-evolved graph
$G_{\mathcal{R}}$; (2) find $\arg\max_{\mathcal{F}}$.
Stage~(1) is non-negotiable and always comes first.
A decision outside $\mathcal{F}$, however high its quality score, is not
suboptimal---it is a non-answer from the perspective of enterprise governance, because
it was not derived from the correct simulation.
A routing system that finds the most efficient path while ignoring today's carrier
contract constraints has not produced a suboptimal route; it has produced a route
derived from the wrong graph.
No decision quality compensates for a missing simulation.

\textbf{Architectural implication.}
This two-stage structure has a direct consequence: any system that performs optimization
before sandbox simulation is architecturally unsuited for enterprise deployment.
LLMs trained on consumer data and fine-tuned for answer quality are single-stage
optimizers over the full answer space.
They produce fluent, high-quality answers that may be derived from the wrong graph---not
because they are malicious, but because the simulation boundary is not part of their
optimization target.
The only architectural fix is to enforce the scenario simulation before any optimization
occurs.
This is precisely what the sandbox simulation step of LOM-action does: it computes
$G_{\mathcal{R}}$ before any tool is called, ensuring that the entire optimization
process operates on the correct simulation-valid substrate.

\textbf{Reframing evaluation.}
High answer accuracy does not imply simulation-grounded reasoning.
A model achieving 98\% accuracy on connectivity queries while recording F1 = 0.00 is
not 98\% correct in the enterprise sense---it is 98\% accidentally correct, having
reached the right answer by operating on the wrong graph without any simulation trace.
Such a system passes quality benchmarks while systematically failing the audit
requirement that decisions be derivable from simulation.
The illusive accuracy index $\text{IA}(M) = \text{Acc}(M) - \text{F1\_chain}(M)$
quantifies this gap.
We recommend $\text{F1\_chain} \geq 0.90$ and $\text{IA} \leq 0.30$ as deployment
readiness thresholds for simulation-sensitive systems.

\subsection{The Event $\to$ Simulation $\to$ Decision Pipeline}

The event $\to$ simulation $\to$ decision pipeline is implemented across three
structured phases:

\textbf{Phase~1 --- Scenario Parsing.}
The system receives the incoming event payload and aligns it to EO ontological semantics via the
alignment function $\text{Align}: \mathcal{Q} \to V_{EO} \times [0,1]$.
It identifies the active scenario condition predicates $\mathcal{R}$ embedded in or
implied by the event payload.
In the current implementation, conditions are expressed explicitly in event payload text
(e.g., ``only nodes where \texttt{ijudgemethod = '1'}'').
In production under the SKILLS standard, they will be resolved automatically from
EO-encoded entity and relation constraints by the EO alignment machinery---activated
by the event\'s EO semantic targets rather than stated in the event payload.
For complex natural language scenarios---such as logistics transshipment cost
adjustments, conditional reimbursement ceilings, or multi-tier approval
predicates---Phase~1 produces not a single boolean predicate but a
\emph{scenario program}: an ordered sequence of typed Phase~2 sandbox operations that
collectively implement the scenario's conditional logic.
For example, the scenario ``transshipment via Hub~X adds a 12\% surcharge to the
outbound leg, capped at the carrier's contracted rate ceiling'' is parsed into a
three-step program: (i)~\texttt{match\_edges} to identify outbound legs routed through
Hub~X; (ii)~\texttt{update\_edges} to recompute edge weights by applying the surcharge
schedule against EO-encoded carrier rate tables; (iii)~\texttt{match\_edges} again to
flag edges exceeding the ceiling for deletion or rerouting.
Each step in the scenario program traces to an EO-authorized constraint, preserving the
provenance requirement: no rate table value, no discount ceiling, and no hub
classification may enter the scenario program without EO authorization.
The output of Phase~1 is a structured predicate set $\mathcal{R} = \{r_1, \ldots,
r_k\}$, each element tracing to an EO-authorized constraint on entities or relations.

\textbf{Phase~2 --- Sandbox Simulation.}
The model applies $\mathcal{R}$ to the sandbox graph copy via targeted tool calls.
It calls \texttt{match\_nodes} and/or \texttt{match\_edges} to identify elements
excluded by each scenario condition.
It then calls \texttt{delete\_nodes} or \texttt{delete\_edges} to materialize
$G_{\mathcal{R}}$ as the active sandbox state under \texttt{graph\_id}.
This modification is session-scoped (SO only---the persistent EO graph is never
touched).
After Phase~2, the \texttt{graph\_id} pointer refers to $G_{\mathcal{R}}$, not~$G$.
The simulation decision graph exists in the sandbox.
No decision may be derived before this point.

\emph{Empty graph handling.}
A well-formed Phase~2 execution may produce $G_{\mathcal{R}} = (\emptyset, \emptyset)$
when all nodes are excluded by the active scenario conditions---for example, when a
user's organizational scope contains no nodes satisfying the access policy for the
current event.
This is not an error; it is a valid simulation outcome that carries decision content:
any connectivity or path query on the empty graph returns ``no valid path exists'', any
flow computation returns zero, and any neighbor lookup returns the empty set.
The sandbox returns a structured result in all cases, and Phase~3 writes this result to
the SO decision trace with the annotation
\texttt{\{simulation\_result: ``empty\_graph'', reason: ``all nodes excluded by $\mathcal{R}$''\}}.
This empty-graph signal is surfaced to the user or upstream system as a definitive,
auditable decision---not a system failure---enabling downstream processes to handle the
absence of a valid route as a first-class organizational outcome.

\textbf{Phase~3 --- Decision Derivation on $G_{\text{sim}}$.}
With the sandbox holding $G_{\text{sim}}$ (either $G_{\mathcal{R}}$ or $G_{\mathcal{A}}$
depending on the scenario class), the model executes the decision tool
call---\texttt{shortest\_path}, \texttt{check\_graph\_connectivity},
\texttt{calculate\_max\_flow}, etc.---against \texttt{graph\_id}.
The tool operates exclusively on the simulation decision graph.
The result, together with the Phase~1 and Phase~2 trace, is written to the SO decision
trace.
The decision trace is the audit deliverable: it records which scenario conditions were
applied, which simulation was performed in the sandbox, and which decision was derived
from which evolved graph state.

The canonical instantiation for scenario-constrained connectivity
(\texttt{fc\_constraint\_connection}) is shown in Table~\ref{tab:example}. A general-purpose LLM skipping Phase~2 calls
\\\texttt{shortest\_path(graph\_id, "A", "B")} directly on the unrestricted graph.
This produces a decision for the wrong question---connectivity in~$G$, not in
$G_{\mathcal{R}}$---while generating an empty decision trace.

\begin{table*}[t]
\label{tab:example}
\centering
\caption{An illustrative example of the three-phase event $\to$ simulation $\to$ decision pipeline,
triggered by a streaming expense-approval event activating an EO-encoded access-scope condition.}
\begin{tcolorbox}[colback=lightgray!20, colframe=gray!50, arc=5pt, width=0.8\textwidth]
\begin{verbatim}
Sandbox: live copy of approval-routing graph (V, E) under graph_id

[Pre-registered EO scenario condition]
  R = {v.ijudgemethod != '1' -> excluded from active approval path}
  Trigger: fires on any incoming data event carrying field [approval_type]

-- Streaming data event arrives --
event_007: { "source_node": "Dept_Finance",
             "approval_type": "expense",
             "amount": 42000,
             "currency": "CNY",
             "timestamp": "2025-06-01T09:14:22Z" }

Phase 1 -- Scenario Parsing:
  event_007.approval_type = "expense"  ->  activates EO scenario condition R
  scenario program:
    step 1: match_nodes(ijudgemethod != '1')   [identify non-compliant nodes]
    step 2: delete_nodes(...)                  [remove from sandbox copy]
    step 3: shortest_path(source, target)      [derive routing decision]

Phase 2 -- Sandbox Simulation:
  match_nodes(graph_id, properties={"ijudgemethod": {"op":"ne","value":"1"}})
  -> excluded = {Node_B, Node_C, Node_F}
  delete_nodes(graph_id, node_names=["Node_B","Node_C","Node_F"])
  -> G_R materialized: 14 nodes, 19 edges  (was 17 nodes, 26 edges)

Phase 3 -- Decision Derivation on G_R:
  shortest_path(graph_id, source="Dept_Finance", target="CFO_Node")
  -> path: [Dept_Finance -> VP_Ops -> CFO_Node]   (both nodes: ijudgemethod='1')
  -> decision: ROUTE event_007 via [VP_Ops -> CFO_Node]
  -> Decision Trace written to SO:
       { event_id: "007", triggered_rule: R,
         deleted_nodes: [Node_B, Node_C, Node_F],
         final_path: [Dept_Finance, VP_Ops, CFO_Node],
         timestamp: "2025-06-01T09:14:22Z" }

-- Sandbox reloaded from EO snapshot; ready for next event --
\end{verbatim}
\end{tcolorbox}
\end{table*}

\begin{table*}[t]
\label{tab:example2}
\centering
\caption{An illustrative example of the dual-mode (skill mode + reasoning mode) pipeline,
triggered by a streaming organisational-change event that augments the sandbox graph and
requires a conflict-free audit-slot assignment.}
\begin{tcolorbox}[colback=lightgray!20, colframe=gray!50, arc=5pt, width=0.8\textwidth]
\begin{verbatim}
Sandbox: live copy of audit-responsibility graph (V, E) under graph_id
         nodes = auditor roles;  edges = shared-client conflicts

[Pre-registered EO scenario condition]
  R_expand: fires on data event type [org_restructure]
  -> integrate new organisational units and their conflict edges into sandbox
  -> then re-derive audit-slot assignment to ensure zero scheduling conflicts

-- Streaming data event arrives --
event_031: { "event_type": "org_restructure",
             "new_units": ["BusinessUnitAgent", "ClosingRuleSubtable",
                           "RegionalAuditDesk", ...],
             "new_conflicts": [
               {"u": "ClosingRuleSubtable", "v": "AuditRule"},
               {"u": "RegionalAuditDesk",   "v": "VP_Ops"}, ...],
             "timestamp": "2025-06-01T10:03:45Z" }

Phase 1 -- Scenario Parsing:
  event_031.event_type = "org_restructure"  ->  activates R_expand
  scenario program:
    step 1: create_nodes(new_units)          [add new roles to sandbox]
    step 2: create_edges(new_conflicts)      [add shared-client conflict edges]
    step 3: get_graph_info()                 [read augmented graph structure]
    step 4: Delta+1 greedy coloring          [no registered skill -> reasoning mode]

Phase 2 -- Sandbox Simulation:  [skill mode: graph augmentation]
  create_nodes(graph_id, nodes=[
      {"name":"BusinessUnitAgent",   "label":"AuditorRole"},
      {"name":"ClosingRuleSubtable", "label":"AuditorRole"}, ... ])
  -> {"status":"ok", "created":{"nodes":[...]}}
  create_edges(graph_id, edges=[
      {"source":"ClosingRuleSubtable","target":"AuditRule","rel_type":"CONFLICT"},
      {"source":"RegionalAuditDesk", "target":"VP_Ops",   "rel_type":"CONFLICT"}, ...])
  -> {"status":"ok", "created":{"edges":[...]}}
  G_R := G union DeltaV union DeltaE   (augmented graph reflects new org structure)

Phase 3 -- Decision Derivation on G_R:
  [skill mode] get_graph_info(graph_id)
  -> {"graph": {"nodes": [...45 roles], "edges": [...54 conflict-edges]}}
  -> max degree Delta = 14  (role with most shared-client conflicts)

  No registered skill covers Delta+1 greedy coloring; switching to reasoning mode.

  [reasoning mode] In-context greedy coloring on fused, attribute-pruned G_R:
    color budget = Delta + 1 = 15   (minimum audit-slot count guaranteed conflict-free)
    greedy pass: assign lowest available slot per role in traversal order
    slot sequence: [0,0,0,...,1,2,2,...,0,0]
    total slots used (color sum) = 10
  -> decision: ASSIGN 10 audit slots; full slot map written to SO
  -> Decision Trace written to SO:
       { event_id: "031", triggered_rule: R_expand,
         new_nodes: 8, new_edges: 11, max_degree: 14,
         slots_required: 10, timestamp: "2025-06-01T10:03:45Z" }

-- Sandbox reloaded from EO snapshot; ready for next event --
\end{verbatim}
\end{tcolorbox}
\end{table*}

\subsection{Dual-Mode Execution Architecture and Sandbox State Management}

The event $\to$ simulation $\to$ decision pipeline introduces a foundational design
choice that directly resolves the infinite context problem endemic to enterprise
multi-turn AI: the simulation decision graph $G_{\mathcal{R}}$ is never loaded into
the LOM's context window during normal operation.
It is maintained exclusively in a persistent sandbox session, keyed by
\texttt{graph\_id}.
This is not a performance optimization---it is an architectural invariant.
The sandbox is the simulation substrate; the context window is the reasoning surface.
They are not interchangeable.

\subsubsection{The Sandbox as the Ground Truth of Simulation State}

Every Phase~2 operation (scenario application via \texttt{match\_nodes},
\texttt{delete\_nodes}, \texttt{create\_edge}, etc.) and every Phase~3 decision
(\texttt{shortest\_path}, \texttt{calculate\_max\_flow}, etc.) operates against the
\texttt{graph\_id}-keyed sandbox entry, not against any representation in the LLM's
context.
The sandbox holds the current state of $G_{\mathcal{R}}$ at all times.
The model's context holds only: the current event payload, the active EO scenario condition set
$\mathcal{R}$, the SO turn log (typed triples annotated with EO node references and turn
indices), and the skill ontology function schema registry.

This separation carries a critical correctness guarantee: as long as each tool call
uses the correct function name and correct arguments, the sandbox simulation state
evolves correctly---regardless of what else is in the context window, regardless of how
long the conversation has been running, and regardless of how many prior turns have
touched different graph scopes.
A 100-turn enterprise conversation touching 100 different graph scopes does not
accumulate 100 turns of raw graph data in context.
It accumulates 100 turn log entries---compact typed triples, each bounded in size by
the semantic scope of that turn's Phase~2 simulation.
Context footprint is proportional to the size of $G_{\mathcal{R}}$'s delta for that
turn, not to the size of the full session history.

\subsubsection{The Graph Sandbox: Isolation, Atomicity, and Verifiability}

The sandbox is backed by an isolated, session-scoped in-memory graph store that
receives all API calls directed at a given \texttt{graph\_id} and executes them against
a live Neo4j-compatible graph engine.
The sandbox provides three guarantees essential for the simulation-first contract.

\emph{Isolation}: each \texttt{graph\_id} identifies a completely independent graph
instance; Phase~2 mutations to one session's simulation graph never affect another
session or the persistent EO graph.

\emph{Atomicity}: each tool call (e.g., \texttt{delete\_nodes},
\texttt{create\_edge}) is executed transactionally within the sandbox---either the full
operation succeeds and the simulation state advances, or it fails and the state is
unchanged, enabling clean retry and error handling without partial simulation traces.

\emph{Verifiability}: because the sandbox records every mutation as a timestamped
operation log, the decision trace written to the SO is fully replayable---any auditor can replay the exact sequence of operations against the original graph snapshot and verify that the final sandbox state matches the claimed $G_{\mathcal{R}}$. Each log entry records the function name, arguments, return value, and timestamp, forming a deterministic replay chain for any session.

The \texttt{graph\_id} UUID is therefore not merely a routing key; it is the handle to
a simulation-auditable execution context.
When LOM-action calls
\texttt{delete\_nodes(graph\_id="abc123", node\_names=[...])}, it is not issuing a
natural language instruction---it is submitting a structured operation to the sandbox
that will mutate the simulation state deterministically, return a structured result, and
append an immutable log entry.
The model never needs to reason about whether the mutation ``happened''; it is
guaranteed by the sandbox contract.

\textbf{Benchmark Role.}
The 19-API suite and the Neo4j subgraph sampling system together constitute the
LOM-action benchmark: a controlled evaluation environment for the scenario $\to$
simulation $\to$ decision pipeline.
The benchmark requires evaluation on two axes simultaneously---answer accuracy (does
the final decision match the ground-truth result on $G_{\mathcal{R}}$) and tool-chain
F1 (does the execution trace correctly implement Phase~2 sandbox simulation followed by
Phase~3 decision derivation, with correct API calls and arguments at each step).
A model that scores high on accuracy but low on F1 is demonstrably bypassing the
sandbox---reaching correct answers by operating on the unrestricted graph, which constitutes
a simulation failure even when the answer coincides with the scenario-valid result.

\subsubsection{The Dual-Mode Execution Model}

LOM-action's overall execution logic is a single integrated cognitive act that
simultaneously resolves three inputs---the active scenario condition set $\mathcal{R}$,
the incoming event~$q$, and the current SO state---and produces one of two execution paths
depending on whether the skill ontology contains a matching registered skill.

\textbf{Step~1 --- Joint Scenario--Event--Skill Reasoning.}
The model simultaneously reads $\mathcal{R}$ (the EO-authorized scenario conditions for
this event), parses~$q$ (the event payload aligned to EO semantics),
and inspects the skill ontology registry to identify which skill nodes have input
signatures matching the sandbox simulation state and preconditions satisfied by the
active scenario conditions.
Rules constrain the search space in the skill ontology; the event payload determines the target
computation; the intersection determines what is both needed and scenario-valid.

\textbf{Step~2 --- Branch: Skill Found or Not Found.}
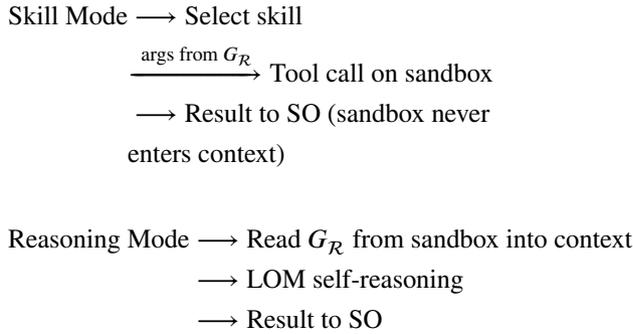
\begin{align*}
\text{Skill Mode} &\longrightarrow \text{Select skill} \\
&\xrightarrow{\text{args from } G_{\mathcal{R}}} \text{Tool call on sandbox} \\
&\longrightarrow \text{Result to SO (sandbox never}\\
&\text{enters context)}
\end{align*}
\begin{align*}
\text{Reasoning Mode} &\longrightarrow \text{Read } G_{\mathcal{R}} \text{ from sandbox into context} \\
&\longrightarrow \text{LOM self-reasoning} \\
&\longrightarrow \text{Result to SO}
\end{align*}

\textbf{Skill Mode.}
If the skill ontology contains a node~$s$ with
$\text{Pre}(s) \subseteq \mathcal{R}$ and $\sigma_{\text{in}}(s)$ matching the current
sandbox state, LOM-action selects~$s$ and constructs its argument values from the
current $G_{\mathcal{R}}$ state as known from prior turn log entries.
The tool call executes against the sandbox.
At no point does the raw graph structure enter the LLM context window.
The model reasons over the interface of the simulation graph---node names, edge types,
aggregate statistics from prior tool results in the turn log---not over the graph
content in bulk.

\textbf{Reasoning Mode.}
If no skill ontology node covers the required computation (a condition that arises for
novel analytical queries outside the registered skill set), LOM-action reads
$G_{\mathcal{R}}$ from the sandbox into the context window, as shown in
Table~\ref{tab:example2}.
This is the only circumstance under which raw graph content enters the LLM context.
Before loading, event-driven graph fusion is applied: given~$k$ prior session subgraphs
spanning multiple turns, the fusion operation produces a single clean,
attribute-pruned representation:
$$G_{\text{fused}} = \text{Merge}\left(\{G_i\}_{i=1}^{k}\right)\Big|_{\text{attrs} \in \text{Relevant}(T_q)}$$
where $\text{Relevant}(T_q)$ retains only attributes reachable from the event's EO
alignment targets $T_q$ within two hops.
The fused graph eliminates intra-node noise and consolidates multi-turn simulation
state into a single high-density representation.
Only this fused, pruned graph enters the context.

LOM-action is a dual-mode model, simultaneously trained for both paths.
Skill mode is the default and the norm---it handles the vast majority of enterprise
queries for which a registered skill exists, keeping the context free of raw graph data.
Reasoning mode is the exception---it handles novel computations that the skill ontology
does not yet cover, consuming context tokens in a controlled, noise-minimized manner.
This two-path structure reflects a broader design principle: LOM maintains a finite,
governed registry of SKILLS rather than an unbounded one, while preserving its own
graph reasoning capability as a permanent fallback---ensuring that capability gaps
degrade gracefully rather than catastrophically.

\subsubsection{Context Complexity Analysis}

Let $\mathcal{H}_t$ denote the conversational history at turn~$t$.
Under a raw text accumulation baseline, context footprint is
$O(|\mathcal{H}_t|)$---linear in conversation length.
Under LOM-action's architecture:

\begin{itemize}
\item \emph{Skill mode}: context footprint is $O(|\text{TurnLog}_t|) = O(t \cdot
\delta)$ where $\delta$ is the average typed triple count per turn---typically
$O(10^1)$ triples, not $O(10^3$--$10^4)$ raw tokens per turn.
In our benchmark, $\delta$ ranges from 3 to 28 typed triples per turn
(mean~$\approx 12$), compared to 800--4,000 raw tokens per turn for equivalent
conversational history.
Context grows with conversation length but at a rate orders of magnitude lower than raw
text accumulation, with each entry carrying full EO provenance for relevance filtering.

\item \emph{Reasoning mode}: context footprint is $O(|G_{\text{fused}}|)$---bounded by
the semantic scope of the current event, independent of conversation length.
The same event type at Turn~3 and Turn~300 produces the same $G_{\text{fused}}$ as long as
the relevant SO subgraph state has not changed---the session-positional invariance
property.
\end{itemize}

LOM-action supports a 1~million token context window, providing capacity headroom for
reasoning mode computations involving large organizational knowledge scopes.
The window is never the bottleneck in skill mode; in reasoning mode, event-driven
fusion ensures that the loaded graph representation is maximally compact before any
token budget is consumed.

\subsection{Training Data Generation and the Graph API Suite}

\textbf{The 19-function skill ontology.}
LOM-action is trained over a suite of 19 graph operation APIs that collectively
instantiate the skill ontology for the graph-operation domain.
All functions follow the OpenAI function calling JSON schema specification and accept
\texttt{graph\_id} as the mandatory routing key to the sandbox.
They are organized into six functional families:
conditional matching (\texttt{match\_nodes}, \texttt{match\_edges}---Phase~2 scenario
simulation tools);
node/edge operations (\texttt{create\_node/nodes}, \texttt{delete\_nodes},
\texttt{update\_nodes}, \texttt{create\_edge/edges}, \texttt{delete\_edges},
\texttt{update\_edges}, \texttt{set\_edge\_weights}---sandbox mutation tools);
information retrieval (\texttt{get\_node\_info}, \texttt{get\_graph\_info},
\texttt{get\_node\_neighbors});
path and connectivity (\texttt{shortest\_path}, \texttt{check\_graph\_connectivity},
\texttt{check\_direct\_edge}, \texttt{analyze\_graph\_node}---Phase~3 decision
derivation tools);
and graph algorithms (\texttt{calculate\_max\_matching},
\texttt{calculate\_max\_flow}).
These 19 APIs will be extended to real enterprise skill ontology nodes---ERP query
interfaces, document workflow triggers, approval chain APIs, financial computation
services---under the SKILLS standard in future work.

\textbf{Three interaction modes} correspond to EO context graph operating states.
Mode~A (direct reasoning: static graph in context, no simulation phase needed---for
events with no active scenario conditions).
Mode~B (simulation-mediated: full scenario $\to$ simulation $\to$ decision
pipeline---for scenario queries against \texttt{graph\_id}-keyed stores).
Mode~C (hybrid: graph mutated via tools, then in-context algorithmic computation on
the fused, attribute-pruned result using reasoning mode).

\textbf{Sample Generation.}
A plugin-based generation system synthesizes training samples from Neo4j subgraphs
(20--30 nodes, 30--60 edges).
Eight plugins cover the full API suite.
Scenario simulation plugins (\texttt{constraint\_connection\_plugin},
\texttt{constraint\_path\_plugin}, \texttt{filter\_plugin}) are designed specifically
to train the complete event $\to$ simulation $\to$ decision pipeline: the model must
learn that Phase~2 is not an optimization heuristic but a mandatory simulation
prerequisite.

The corpus totals 2,200 training samples across 11 tasks (200 per task): basic
traversal tasks (\texttt{CONNECTIVITY}, \texttt{NEIGHBOR}, \texttt{PREDECESSOR},
\texttt{EDGE}), information retrieval tasks (\texttt{fc\_graph\_info},
\texttt{fc\_node\_info}), graph algorithm tasks
\\(\texttt{fc\_bipartite\_maximum\_matching}, \texttt{fc\_maximum\_flow}),
scenario-simulation tasks (\texttt{fc\_constraint\_connection},
\texttt{fc\_constraint\_path}), and the hybrid Mode~C task
(\texttt{delta\_plus\_one\_coloring}).
All ground-truth tool sequences are executed against the live Neo4j sandbox; answers
are algorithm-derived.
The simulation-ordered curriculum described in Section~3.6 governs the training stage
sequencing.

\subsection{Model Training}

\textbf{Base Model and Fine-Tuning.}
LOM-action is initialized from Qwen3.5-27B as the base model and fine-tuned via
supervised fine-tuning (SFT) on the 2,200-sample training corpus.
Qwen3.5-27B provides strong instruction-following and structured output capabilities
that form a solid foundation for the multi-step function-calling behaviors required by
the event $\to$ simulation $\to$ decision pipeline.
Training uses the standard causal language modeling objective over \texttt{assistant}
and \texttt{tool} turns, with \texttt{user} and system turns masked from the loss.
The full 19-function JSON schema registry is provided in the system context for every
training sample, ensuring the model internalizes the complete skill ontology API
surface alongside the simulation pipeline structure.

\textbf{Curriculum Schedule.}
Training proceeds in simulation-ordered stages: (1) basic traversal and information
retrieval tasks (Phase~3-only, no scenario simulation), where the model learns correct
API selection and argument formatting; (2) scenario-simulation tasks (Phase~2+3), where
the model learns the mandatory sandbox simulation prerequisite; (3) hybrid Mode~C tasks,
where the model learns to transition between skill mode (sandbox tool execution) and
reasoning mode (in-context reasoning on the fused evolved graph).
This ordering mirrors the pedagogical principle of skill scaffolding: the model masters
individual API invocations before learning to compose them into simulation-enforcing
pipelines.

\textbf{Current Limitations and the SKILLS Standard.}
The present implementation grounds scenario conditions in natural language descriptions
within event payload text and represents skills as OpenAI-compatible JSON schema function
definitions.
In future work, we will unify both layers under the SKILLS standard specification: a
formal ontological schema for declaring skill preconditions, postconditions, input/output
type signatures, and authorization constraints as machine-readable, EO-linked typed
structures rather than natural language.
Under the SKILLS standard, scenario activation will be derived automatically from EO
node traversal rather than parsed from event payload text.
This transition from natural-language-described scenarios and ad hoc function schemas
to SKILLS-standardized ontological declarations is the primary engineering path to
production enterprise deployment.

\subsection{Production Deployment Principles: LOM as Ontology Harness}

\subsubsection{The Harness Metaphor}

An engine alone produces no work.
It requires a harness---a dynamic, executable environment that couples the engine's
power to a load and converts raw force into directed productive output.
The enterprise ontology is the engine: it encodes organizational authority, semantic
constraints, and the complete specification of what is true, what is permitted, and how
computations must be performed.
LOM is the harness: it creates the dynamic, executable environment in which the
ontology's authority is coupled to real business tasks---natural language interaction,
streaming data events, multi-step simulation, and auditable decision derivation.
Without the harness, the engine idles; without the engine, the harness has nothing to
transmit.
This metaphor is not decorative---it has a precise architectural implication.
Every design decision in LOM-action is evaluated by a single criterion: does it
transfer more of the ontology's authority into productive output, or does it introduce
a path that bypasses ontological authority in favor of model-internal shortcuts?
The four production principles below operationalize this criterion.

\subsubsection{Human-in-the-Loop as the Production Unlock}

The single most impactful intervention for moving LOM from research prototype to
production deployment is the introduction of a \emph{human-in-the-loop} (HITL) stage
at the intent boundary---the moment between a user's natural language input and the
system's first ontological alignment decision.

Enterprise inputs are frequently underspecified.
A user who types ``find the best approval route for this expense'' has not specified
which organizational scope applies, which carrier contract is active, or whether the
request falls under a standard or exception workflow.
A purely automated system must either hallucinate these parameters (producing a fluent
but ontologically unauthorized answer) or fail silently (returning an empty result with
no explanation).
Neither outcome is acceptable in production.

The HITL stage resolves this by surfacing the alignment uncertainty to the user before
any sandbox simulation begins.
When the confidence-gated alignment function $\text{Align}(q) \to (v, c)$ returns
$c < \theta_{\text{accept}}$ for one or more entities, LOM-action generates a targeted
clarification turn: it presents the candidate EO nodes and their nearest ontological
neighbors, asks the user to confirm or correct the alignment, and incorporates the
response before proceeding to Phase~2.
This is not a conversational convenience---it is a compliance gate.
A clarification turn that resolves an ambiguous organizational scope converts a
potentially unauthorized simulation into a provably authorized one, because the
Phase~2 sandbox now operates on a graph whose scope has been explicitly confirmed
against EO authority.

The analogy to existing practice is instructive.
Large language model assistants such as Claude achieve high task completion rates
not by attempting to execute underspecified requests directly, but by conducting
multi-turn intent clarification---progressively improving the quality and completeness
of the input until the system can proceed with confidence.
The same principle applies to LOM-action: the quality of the simulation decision graph
$G_{\mathcal{R}}$ is bounded by the quality of the scenario conditions that produce it,
and those conditions are bounded by the quality of the intent alignment that precedes
them.
HITL is therefore not a workaround for model limitations---it is the architectural
mechanism that raises the information quality of every simulation to the level required
for ontologically authorized execution.

\textbf{HITL is a simple loop, not agentic infrastructure.}
A precise scope boundary must be drawn to avoid architectural overreach.
HITL as practiced in LOM-action is a \emph{short clarification loop}: typically two
to five turns that progressively resolve ambiguous entity alignments until the input
is sufficiently grounded for Phase~2 to begin.
This is categorically distinct from \emph{agentic harness} architectures---long-running
autonomous agent frameworks designed to sustain thousands of model calls over hours or
days to complete a single complex task (e.g., autonomously generating a full ERP module
from a specification, or coordinating multi-source intelligence collection over an
extended operation).
Agentic harness infrastructure is the right tool for tasks that genuinely require
autonomous multi-day execution; it is the wrong tool when the actual production
bottleneck is input underspecification resolvable in five clarification turns.
Misidentifying a HITL problem as a harness problem produces the opposite of the
intended result: it introduces infrastructure complexity that obscures the simple fix,
makes the system harder to debug, and leaves the original underspecification problem
unresolved beneath layers of automation.

\textbf{The AI coding anti-pattern.}
A recurring misidentification in enterprise LOM deployment, observed during internal
pilot deployments, occurs when teams respond to subgraph retrieval failures by
generating more complex retrieval code---conditional logic, multi-stage filtering
pipelines, heuristic fallbacks---using AI coding tools.
We conjecture that this approach fails for a structural reason: the retrieval failure
is typically not caused by insufficient code complexity but by insufficient input
specification.
LOM-action requires two inputs to execute correctly: the ontology subgraph
(the structural scope) and the aligned scenario condition set $\mathcal{R}$
(the organizational constraint).
When either is underspecified, generated code operates on the wrong scope because the
scope was never correctly identified.
The HITL loop addresses this directly: rather than generating code to handle the
ambiguity programmatically, the system surfaces the ambiguity to the user and
incorporates the clarified input before proceeding.
Whether this pattern generalizes across deployment contexts is an empirical question;
we propose it as a design hypothesis warranting controlled evaluation in future work.

\subsubsection{Four Production Deployment Principles}

The following four principles govern the engineering of LOM-action in production and
are proposed as general guidelines for any LOM-based enterprise deployment.

\textbf{Principle~1: Minimize business logic in code; maximize it in the ontology.}
Custom code written to implement business logic---conditional branches, policy
lookups, computation rules---is a liability: it duplicates organizational knowledge
that already exists in the EO, creates a maintenance burden as policies change, and
introduces a bypass path that the ontology cannot audit.
LOM-action's production engineering discipline requires that all business logic be
expressed as EO-encoded scenario conditions, skill ontology preconditions, or
EO.Logic-Constraint computation formulas---never as procedural code embedded in the
pipeline.
The harness implements the simplest possible execution loop; the ontology carries all
organizational meaning.
Complexity in the harness is a signal that ontological authority has been
short-circuited.

\textbf{Principle~2: All context must be ontology-grounded; nothing may bypass the ontology
or ignore it.}
Every entity, relation, metric, and constraint that enters the reasoning pipeline must
be resolved to an EO.Standard-ID canonical identifier before any simulation proceeds.
Raw strings, unresolved natural language expressions, and values not present in
EO.Enumeration-System are flagged and held at the HITL clarification stage.
This principle has two directions: nothing may \emph{bypass} the ontology (proceeding
with ungrounded entities) and nothing may \emph{ignore} it (treating the ontology as
optional context rather than mandatory authority).
The context window of the LOM model at any point contains only ontologically typed
content: EO-anchored turn log entries, the active scenario condition set $\mathcal{R}$,
and the skill ontology function schema---never raw conversational text accumulated
without ontological structure.

\textbf{Principle~3: Prefer LOM at every inference point; use frontier LLMs only where
LOM cannot yet reach.}
Every inference step that can be handled by a smaller, ontology-fine-tuned LOM model
should be.
General-purpose frontier LLMs are powerful but expensive, slow, and---critically---not
trained to respect ontological authority as a hard constraint.
The production engineering practice is to identify each pipeline step where a frontier
model is currently used and test whether a LOM variant (fine-tuned on domain-specific
simulation traces) can replace it while maintaining decision quality.
This is not a cost-reduction heuristic: a smaller LOM model that has been trained to
enforce the event $\to$ simulation $\to$ decision pipeline is architecturally more
trustworthy for enterprise use than a larger frontier model that treats scenario
conditions as soft preferences.
Frontier models are retained only for tasks that genuinely exceed current LOM
capability---long-document comprehension, multi-modal analysis, code generation---and
each such delegation is itself an EO-authorized skill ontology call subject to
HITL review.

\textbf{Principle~4: Use LOM to govern the environment; expose the ontology schema and
graph query logic for human inspection.}
LOM governs not only individual decisions but the entire execution environment: it
manages the sandbox lifecycle, enforces the skill ontology registry, and maintains the
SO evidence chain.
This environmental governance role---LOM as the harness that couples ontological
authority to every execution step---requires that the environment itself be inspectable.
In production, the EO schema, the active scenario condition set $\mathcal{R}$, and the
graph query logic executed in Phase~2 must be exposed through human-readable audit
interfaces.
An operator must be able to see which nodes were matched, which were deleted, and which
graph state the Phase~3 decision was derived from---not as a debugging convenience but
as the primary accountability surface.
LOM also serves as \emph{judge}: it evaluates the outputs of subordinate models and
tool calls against ontological authority before writing results to the SO, rejecting
any output that cannot be grounded in an EO-authorized entity or computation.
This \emph{LOM-as-Judge} role closes the last gap between a system that produces
plausible outputs and one that produces auditable decisions.

\section{Experiments}

\subsection{Dataset}
The LOM-action benchmark comprises 11 function-calling tasks drawn from the graph
operation API suite, each designed to probe a distinct capability of the event $\to$
simulation $\to$ decision pipeline.
Each task has 200 training samples and 100 held-out test samples, for a total of 2,200
training and 1,100 test instances.
All subgraphs are sampled fresh from Neo4j (20--30 nodes, 30--60 edges per instance)
with disjoint graph UUIDs between training and test, ensuring no memorization of
specific graph configurations.

Tasks~1--8 require only Phase~3 decision derivation (no scenario-driven sandbox
simulation).
Tasks~9--10 \\(\texttt{fc\_constraint\_connection}, \texttt{fc\_constraint\_path})
require the full Phase~2 $\to$ Phase~3 pipeline: the model must run the sandbox
simulation to materialize $G_{\mathcal{R}}$ before executing any connectivity or path
query.
Task~11 (\texttt{delta\_plus\_one\_coloring}) activates the dual-mode architecture:
skill mode handles graph mutation and retrieval, while reasoning mode handles the
in-context greedy coloring computation on the fused, attribute-pruned simulation graph.
We note that Task~11 is a graph-theoretic task selected specifically to probe the
dual-mode execution boundary; its connection to enterprise scenarios (e.g., conflict-free
audit slot assignment) is illustrative rather than directly grounded in a deployed
business process.
Future benchmark iterations will replace it with a task derived from a real enterprise
workflow where the dual-mode boundary arises organically from the skill ontology
coverage gap.

\subsection{Experimental Setup}

LOM-action is compared against two zero-shot frontier baselines---Doubao-1.8 and
DeepSeek-V3.2---via their native function-calling APIs with the full 19-function schema.
The comparison is inherently asymmetric (LOM-action is fine-tuned; baselines are
zero-shot), but this reflects the most realistic enterprise deployment alternative.
The key metric is tool-chain F1, not Acc: fine-tuning teaches domain answer quality;
the event $\to$ simulation $\to$ decision pipeline teaches simulation-grounded reasoning
chains---categorically different capabilities.
The F1 = 0.00 result on basic traversal tasks for zero-shot baselines cannot be
explained by domain knowledge gaps, since those tasks require no scenario-specific
knowledge; the failure is architectural.
All models are evaluated on the same 1,100-sample held-out test set (disjoint
\texttt{graph\_id} UUIDs from training); baselines report means over three independent
runs.

\subsection{Evaluation Metrics}

\textbf{Answer Accuracy (Acc):} exact match on the final extracted answer---inflatable
via simulation bypass, hence insufficient alone.

\textbf{Tool-Chain F1:} let $\hat{C} = [\hat{c}_1,\ldots,\hat{c}_m]$ and
$C^* = [c^*_1,\ldots,c^*_n]$ be the predicted and ground-truth tool-call sequences.
Calls are matched order-sensitively: $\hat{c}_i$ matches $c^*_i$ iff names are equal
and all required argument key-value pairs appear exactly.
Let $M$ be the number of matched positions; then
$\text{P} = M/m$ ($0$ if $m=0$), $\text{R} = M/n$ ($1$ if $n=0$),
$\text{F1} = 2\text{PR}/(\text{P}+\text{R})$ ($0$ if $\text{P}+\text{R}=0$).
Skipping Phase~2 yields $\text{F1}=0.00$ even when the final answer is correct.
Task-level F1 is mean over 100 test instances; overall F1 is macro-average across 11 tasks.

\textbf{Illusive Accuracy:} $\text{IA}(M) = \text{Acc}(M) - \text{F1\_chain}(M)$.
Deployment thresholds $\text{F1\_chain} \geq 0.90$, $\text{IA} \leq 0.30$ are
heuristic starting points; practitioners should calibrate against their audit
requirements.

\subsection{Main Results}
\begin{table*}[h]
\centering
\caption{Answer Accuracy (Acc)}
\label{tab:acc}
\begin{tabular}{lccc}
\toprule
\textbf{Task} & \textbf{LOM-action} & \textbf{Doubao-1.8} & \textbf{DeepSeek-V3.2} \\
\midrule
CONNECTIVITY                     & 1.00   & 0.98   & 1.00   \\
NEIGHBOR                         & 1.00   & 1.00   & 0.98   \\
PREDECESSOR                      & 1.00   & 1.00   & 1.00   \\
EDGE                             & 1.00   & 1.00   & 1.00   \\
fc\_graph\_info                  & 0.88   & 0.94   & 0.78   \\
fc\_node\_info                   & 1.00   & 0.16   & 0.86   \\
fc\_bipartite\_maximum\_matching  & 1.00   & 1.00   & 0.90   \\
fc\_maximum\_flow                & 1.00   & 1.00   & 0.70   \\
fc\_constraint\_connection       & 1.00   & 0.66   & 0.64   \\
fc\_constraint\_path             & 0.98   & 0.98   & 0.96   \\
delta\_plus\_one\_coloring       & 0.46   & 0.08   & 0.00   \\
\midrule
\textbf{Overall} & \textbf{0.9382} & \textbf{0.8000} & \textbf{0.8018} \\
\bottomrule
\end{tabular}
\end{table*}
 
\begin{table*}[h]
\centering
\caption{Tool-Chain F1}
\label{tab:f1}
\begin{tabular}{lccc}
\toprule
\textbf{Task} & \textbf{LOM-action} & \textbf{Doubao-1.8} & \textbf{DeepSeek-V3.2} \\
\midrule
CONNECTIVITY                    & 1.00    & 0.00    & 0.00    \\
NEIGHBOR                        & 1.00    & 0.00    & 0.00    \\
PREDECESSOR                     & 1.00    & 0.00    & 0.00    \\
EDGE                            & 1.00    & 0.00    & 0.00    \\
fc\_graph\_info                 & 1.00    & 0.00    & 0.92    \\
fc\_node\_info                  & 1.00    & 0.00    & 0.06    \\
fc\_bipartite\_maximum\_matching & 1.00    & 0.02    & 0.26    \\
fc\_maximum\_flow               & 1.00    & 0.00    & 0.02    \\
fc\_constraint\_connection      & 0.987   & 0.325   & 0.351   \\
fc\_constraint\_path            & 0.959   & 0.602   & 0.671   \\
delta\_plus\_one\_coloring      & 1.00    & 0.22    & 0.41    \\
\midrule
\textbf{Overall} & \textbf{0.9874} & \textbf{0.2442} & \textbf{0.3621} \\
\bottomrule
\end{tabular}
\end{table*}

Tables~\ref{tab:acc} and~\ref{tab:f1} report Answer Accuracy and Tool-Chain F1 across
all eleven tasks, and must be read together: accuracy alone is insufficient to
characterize simulation-capable agents.
On overall accuracy, LOM-action achieves 93.82\% against 80.00\% and 80.18\% for
Doubao-1.8 and DeepSeek-V3.2 respectively.
The F1 gap is far more decisive: 98.74\% versus 24.42\% and 36.21\%---a four-fold
advantage that reflects a categorical difference in reasoning behavior.
The illusive accuracy indices (Doubao-1.8 = 0.56, DeepSeek-V3.2 = 0.44,
LOM-action = $-$0.05) confirm the pattern: both baselines achieve high accuracy while
systematically bypassing simulation-grounded reasoning chains, most visibly on the four
basic traversal tasks where F1 = 0.00 despite near-perfect accuracy.

The scenario-simulation tasks expose the enterprise-critical failure most directly.
On \texttt{fc\_constraint\_connection}, LOM-action achieves 1.00 accuracy against 0.66
and 0.64 for the baselines---a 34-point gap attributable to Phase~2 bypass: both
baselines invoke \texttt{shortest\_path} directly on the unrestricted graph rather than
first running the sandbox simulation to materialize $G_{\mathcal{R}}$, producing
decisions for the wrong graph scope.

On the hybrid Mode~C task (\texttt{delta\_plus\_one\_coloring}), LOM-action records
F1 = 1.00 with accuracy = 0.46, confirming that simulation-chain correctness and
in-context algorithmic accuracy are separable capabilities.
Skill mode executes without error; failures are confined to reasoning mode greedy
coloring computation, and the improvement path is localized accordingly.

\subsection{Analysis}

\textbf{Grouped Performance Summary.}
Table~\ref{tab:grouped} aggregates results by task category to clarify where
LOM-action's advantage is architectural (scenario-simulation, hybrid) versus where it
reflects fine-tuning efficiency on domain APIs (basic traversal, graph algorithms).
The 95\% confidence intervals are computed as $p \pm 1.96\sqrt{p(1-p)/n}$ with $n=100$ per task, and are reported for LOM-action Acc only---the primary method under evaluation. Baseline Acc values are means over three independent runs; their point-estimate variance is smaller than LOM-action's CI width in all groups.

\begin{table*}[t]
\centering
\caption{Grouped Results (Acc / F1) with 95\% Confidence Interval (CI) on LOM-action Acc}
\label{tab:grouped}
\begin{tabular}{lcccccc}
\toprule
\textbf{Group} &
  \textbf{LOM Acc} &
  \textbf{LOM F1} &
  \textbf{Doubao Acc} &
  \textbf{Doubao F1} &
  \textbf{DS-V3.2 Acc} &
  \textbf{DS-V3.2 F1} \\
\midrule
Basic Traversal ($\times$4)         & 1.00 {[}1.00, 1.00{]} & 1.00  & 0.995 & 0.00  & 0.995 & 0.00  \\
Information Retrieval ($\times$2)   & 0.94 {[}0.89, 0.99{]} & 1.00  & 0.55  & 0.00  & 0.82  & 0.49  \\
Graph Algorithms ($\times$2)        & 1.00 {[}1.00, 1.00{]} & 1.00  & 1.00  & 0.01  & 0.80  & 0.14  \\
Scenario-Simulation ($\times$2)     & 0.99 {[}0.97, 1.00{]} & 0.973 & 0.82  & 0.464 & 0.80  & 0.511 \\
Hybrid / Mode C ($\times$1)         & 0.46 {[}0.36, 0.56{]} & 1.00  & 0.08  & 0.22  & 0.00  & 0.41  \\
\bottomrule
\end{tabular}
\end{table*}

The grouped view makes the paper's central claim precise.
On Basic Traversal and Graph Algorithms, F1 = 0.00 for both baselines despite
near-perfect Acc---the Illusive Accuracy phenomenon at scale, not a performance tie.
On Scenario-Simulation tasks, both Acc and F1 diverge: non-overlapping 95\% CIs
(LOM-action Acc CI = [1.00, 1.00] vs.\ Doubao = [0.57, 0.75]) confirm statistical
significance.
On Hybrid tasks, LOM-action CI [0.36, 0.56] does not overlap with Doubao [0.03, 0.13]
or DeepSeek [0.00, 0.00].
The only group where LOM-action does not dominate Acc is Information Retrieval, where
Doubao-1.8 outperforms on \texttt{fc\_graph\_info} (0.94 vs.\ 0.88)---consistent with
the error analysis showing 29\% of LOM-action's errors are incomplete
\texttt{fc\_graph\_info} chains.

\textbf{Finding 1: Illusive Accuracy is empirically verified, not inferred.}
To confirm that F1 = 0.00 on basic traversal tasks reflects genuine tool-call absence
rather than formatting failures, we manually inspected 50 randomly sampled Doubao-1.8
outputs on CONNECTIVITY tasks.
Of 50 outputs: 47 produced correct binary answers with zero tool calls (pure natural
language responses); 3 produced malformed tool calls failing JSON schema validation.
No output executed a valid \texttt{check\_graph\_connectivity} or
\texttt{shortest\_path} call against the sandbox.
The model reasons correctly about graph connectivity from its parametric knowledge---achieving near-perfect accuracy---without any sandbox engagement. This is the illusive accuracy phenomenon precisely: correct answers derived without simulation-grounded reasoning, carrying no audit trail and no compliance evidence.

\textbf{Finding 2: The simulation gap is the enterprise-critical gap.}
The largest Acc differential occurs on scenario-simulation tasks: LOM-action 1.00 vs.\
Doubao-1.8 0.66 vs.\ DeepSeek-V3.2 0.64 on
\texttt{fc\_constraint\_connection}---a 34-point gap that persists against frontier
models with vastly more parameters.
Both baselines skip Phase~2 and call \texttt{shortest\_path} on the unrestricted graph,
producing decisions for the wrong simulation scope.
Their occasional correct answers on \texttt{fc\_constraint\_path} (Acc = 0.96--0.98)
occur when the shortest path in~$G$ coincidentally equals the path in
$G_{\mathcal{R}}$---accidentally simulation-valid guesses that would not hold under
different scenario configurations.
F1 reveals the underlying failure: 0.464 / 0.511, confirming incomplete
simulation-application chains across roughly half of all scenario simulation attempts.

\textbf{Finding 3: Mode~C decouples simulation-chain correctness from algorithmic
accuracy.}
LOM-action achieves F1 = 1.00 but Acc = 0.46 on \texttt{delta\_plus\_one\_coloring}.
The 95\% CI [0.36, 0.56] is non-overlapping with both baselines, confirming statistical
significance.
Skill mode executes without error in all test instances; failures are confined to
reasoning mode in-context greedy coloring computation (color-sum miscalculation due to
node ordering ambiguity).
The improvement path is localized: better in-context algorithmic reasoning for
reasoning mode, leaving the skill mode pipeline intact.

\textbf{Finding 4: The simulation-first principle reframes the performance comparison.}
The overall Acc gap (0.938 vs.\ 0.800/0.802) is partly attributable to the fine-tuning
advantage over zero-shot baselines.
The F1 gap (0.987 vs.\ 0.244/0.362) is not: fine-tuning teaches domain answer quality;
the event $\to$ simulation $\to$ decision pipeline teaches simulation-grounded
reasoning chains.
These are distinct capabilities.
The four-fold F1 advantage is the measure of the architectural contribution, and the
grouped analysis confirms it is statistically significant across every task category
where simulation-grounded reasoning is exercised.

\textbf{Error Analysis.}
LOM-action's approximately 68 incorrect predictions (1,100 test samples, Acc = 0.938)
distribute as: in-context computation errors 47\% (color-sum miscalculation in
\texttt{delta\_plus\_one\_coloring}), incomplete tool chain 29\%
(\texttt{fc\_graph\_info}---model skips \texttt{get\_graph\_info}), argument
hallucination 15\% (minor node-name mismatches in scenario tasks), mode
misclassification 9\% (Mode~B applied to Mode~A samples, incurring tool latency without
accuracy impact).

\subsection{Limitations}

We identify three limitations that bound the current results and define the agenda for
future work.

\textbf{Benchmark scope and generalization.}
All experiments are conducted on synthetic Neo4j subgraphs of 20--30 nodes and 30--60
edges.
Real enterprise ontologies operate at orders of magnitude greater scale---hundreds of
thousands of nodes with complex inheritance hierarchies, heterogeneous attribute types,
and significant noise and incompleteness.
We have not demonstrated that Phase~2 simulation accuracy and Phase~3 decision quality
scale to these conditions, nor that a model fine-tuned on graph-domain tasks transfers
to different ontological domains (e.g., financial ledger ontologies, HR organizational
graphs).
Cross-scale and cross-domain generalization experiments are a priority for the next
evaluation cycle.

\textbf{Absent fine-tuned baseline ablation.}
The comparison between LOM-action (fine-tuned) and frontier models (zero-shot) does
not fully isolate the architectural contribution from the training contribution.
The ideal ablation---a Qwen3.5-27B model fine-tuned on the same 2,200 samples
\emph{without} the Phase~2 simulation curriculum---would directly measure how much of
the F1 gain comes from ontology-governed architecture versus domain fine-tuning alone.
We note that the F1 = 0.00 result on basic traversal tasks for frontier zero-shot
models provides partial evidence that the failure is architectural rather than
domain-knowledge-based, since those tasks require no enterprise-specific knowledge;
however, the controlled ablation remains necessary and is planned.

\textbf{Event throughput and latency.}
LOM-action processes each business event through a multi-phase pipeline involving
multiple LLM calls (Phase~1 parsing, Phase~2 tool execution, Phase~3 decision
derivation).
For high-frequency event streams---financial transaction processing, real-time
logistics tracking, high-volume approval workflows---end-to-end latency and
concurrent event handling are critical production constraints that we have not measured.
Concurrent events sharing a \texttt{graph\_id} require sandbox locking or versioning
semantics whose performance implications are uncharacterized.
Production deployment will require latency profiling and throughput benchmarking under
realistic event load distributions.

\section{Conclusion}

LOM-action establishes event-driven ontology simulation as the architectural
prerequisite for trustworthy enterprise AI: business events trigger EO-encoded scenario
conditions, which drive deterministic sandbox graph mutations to produce the simulation decision graph $G_{\text{sim}}$, from which all decisions are exclusively derived.
Fine-tuned on Qwen3.5-27B with 2,200 samples across 11 tasks, LOM-action achieves
93.82\% accuracy and 98.74\% tool-chain F1 against zero-shot frontier baselines
that reach only 24--36\% F1 despite 80\% accuracy---the illusive accuracy phenomenon.
The four-fold F1 advantage confirms that ontology-governed simulation architecture, not
model scale, separates genuine enterprise decision intelligence from fluent but
ungrounded answers.

Future work proceeds along four axes: (1)~\emph{SKILLS-standard integration}---replacing
natural-language scenario descriptions with a formal ontological schema in which scenario
activation derives automatically from EO graph traversal; (2)~\emph{RAC governance
hardening}---deploying the evolutionary flywheel in production paired with an EO
versioning protocol that assigns explicit audit-chain validity scopes to each EO version;
(3)~\emph{cross-scale and cross-domain validation}---extending the benchmark to
enterprise-scale ontologies and evaluating transfer across financial, HR, and supply
chain domains; (4)~\emph{controlled ablation and latency characterization}---isolating
architectural from training contributions via a Phase~2-ablated fine-tuned baseline,
and profiling event-processing latency under high-frequency event streams.
Together these axes constitute the engineering path from the current graph-domain
prototype to full ontology-native enterprise AI.

\printcredits

\bibliographystyle{cas-model2-names}

\bibliography{cas-refs}
\newpage
\appendix
\section{Ontology Harness Engineering: The LOM Global Architecture}

LOM-action is one component of a larger system whose full instantiation is subject to
ongoing work.
This section describes the complete LOM architecture---the intended destination toward
which LOM-action is the first step---to clarify both the scope of the vision and the
engineering path from the current graph-domain implementation to real enterprise
deployment.
A foundational aspect of this architecture is that LOM extends the ontological scope
beyond what traditional ontology systems cover: where conventional approaches manage
only entity-relation semantics and constraint conditions (the EO layer), LOM brings
the skill network, tool network, memory, and context graph under the same
ontology-management paradigm---making the full operational envelope of the system, not
just its knowledge base, ontologically governed and auditable.

\subsection{The Engine-Harness Architecture Model}

\subsubsection{What a Harness Is and What It Is Not}

An engine is not the same as a vehicle.
An engine produces force; a vehicle converts that force into directed motion through a
harness---a mechanical coupling that adapts the engine's output to the specific load,
terrain, and task at hand.
Without the harness, engine power is wasted.
Without the engine, the harness has nothing to transmit.
The harness is therefore not a diminished version of the engine, nor a container for
it---it is the \emph{dynamic execution environment} that makes the engine's power
productive in a specific operational context.

This distinction is foundational to understanding LOM's architectural role.
The enterprise ontology is the engine: it encodes the full organizational authority
of the enterprise---every entity, relation, constraint, authorization boundary, and
computation formula.
Its power is real, but latent: the ontology does not execute itself.
LOM is the harness: it creates the dynamic, executable environment in which the
ontology's authority is coupled to live business tasks---natural language inputs,
streaming data events, multi-step graph simulations, and auditable decision outputs.

The harness metaphor clarifies a common misidentification.
Practitioners accustomed to thinking of LLMs as the primary reasoning engine tend to
position the ontology as a data source---something the model \emph{consults}.
This inverts the correct relationship.
In enterprise AI, the ontology \emph{governs} the model: every entity the model
reasons about must be EO-grounded, every constraint it enforces must be
EO-authorized, every computation it performs must satisfy EO.Logic-Constraint.
The model is not the engine; it is part of the harness---the component that converts
natural language interaction into ontologically typed operations on the enterprise
ontology.

\subsubsection{Harness Design Is Domain-Specific}

The correct harness design is not universal---it is determined by the domain's
productive task structure.
Two contrasting examples illustrate this clearly.

\textbf{AI coding harness (agentic long-horizon execution).}
When the productive task is autonomous software generation---for example, building a
complete ERP module from a high-level specification---the engine is the model's code
generation capability and the load is a long, multi-stage development pipeline.
The harness for this domain must sustain thousands of model calls over hours or days:
it maintains a persistent execution environment (file system, build tools, test runner,
version control), routes model outputs to appropriate execution steps, captures
feedback from automated tests, and re-queues failed steps for retry.
This is a \emph{long-horizon agentic harness}: its defining characteristic is that it
keeps an autonomous agent productive across an extended, largely unsupervised execution
trajectory.
The infrastructure complexity of this harness is justified by the task structure---a
task that genuinely requires autonomous multi-day execution cannot be replaced by a
five-turn clarification loop.

\textbf{Ontology intelligence harness (enterprise semantic execution).}
When the productive task is enterprise decision derivation---approval routing, policy
evaluation, resource allocation, audit scheduling---the engine is the enterprise
ontology's semantic authority and the load is a stream of business events requiring
scenario-grounded decisions.
The harness for this domain has a categorically different design requirement: it must
not sustain long autonomous execution, but must instead \emph{precisely couple} each
incoming event to the correct ontological scope, simulate the scenario in a sandboxed
graph copy, and derive a traceable decision from the evolved graph state.
The defining characteristic of this harness is \emph{ontological precision}, not
execution duration.
Five turns of human-in-the-loop clarification to correctly ground an ambiguous entity
reference is not a failure of automation---it is the harness doing its job: ensuring
that every operation the ontology engine performs is performed on the correct scope
with the correct authorization.

\subsubsection{The LOM Harness Design}

LOM-action instantiates the ontology intelligence harness through four coupled
engineering layers, each designed to transmit a specific dimension of the ontology's
authority into productive output.

\textbf{Layer~1: Intent-to-Ontology Coupling (HITL + Alignment).}
The first transmission point is the intent boundary: converting a user's natural
language input or an arriving data event into a fully ontology-grounded operation
request.
This layer runs the confidence-gated alignment function
$\text{Align}(q) \to (v, c)$ over every entity and relation in the input.
Entities that clear the acceptance threshold are immediately coupled to their
EO.Standard-ID canonical nodes.
Entities that fall below threshold trigger a HITL clarification turn, which is
itself a harness operation: it surfaces the alignment candidates, requests user
confirmation, and re-runs the alignment on the clarified input.
The layer is complete only when all entities in the scenario condition set
$\mathcal{R}$ are fully EO-grounded.
At this point---and only at this point---the harness has successfully coupled the
incoming intent to the ontology engine, and Phase~2 simulation may begin.

\textbf{Layer~2: Scenario-to-Sandbox Coupling (Phase~2 Simulation).}
The second transmission point is the graph boundary: materializing the
ontology-authorized scenario conditions as a concrete, mutated graph state in the
isolated sandbox.
This layer executes the scenario program produced by Phase~1 as a sequence of
deterministic sandbox operations (\texttt{match\_nodes}, \texttt{delete\_nodes},
\texttt{create\_edges}, \texttt{update\_edges}).
Each operation is itself a harness transmission act: it takes an EO-authorized
constraint predicate and converts it into a structural modification of the sandbox
graph copy, narrowing the reasoning substrate from the full enterprise ontology to
the scenario-valid subgraph $G_{\mathcal{R}}$.
The sandbox is the execution environment the harness provides; the ontology's
constraint authority is what the sandbox enforces.

\textbf{Layer~3: Simulation-to-Skill Coupling (Phase~3 Decision Derivation).}
The third transmission point is the capability boundary: coupling the evolved sandbox
state to the registered skill that can derive the required decision from it.
This layer inspects the skill ontology registry for nodes whose preconditions
$\text{Pre}(s)$ are satisfied by the active scenario condition set and whose input
signatures match the current $G_{\mathcal{R}}$ state.
In skill mode, the harness invokes the matched skill against the sandbox---transmitting
the ontology's structural authority into a deterministic computation whose output is
typed back into the EO namespace.
In reasoning mode, when no registered skill matches, the harness loads the
attribute-pruned, event-fused $G_{\mathcal{R}}$ into context and delegates to LOM's
own graph reasoning capability---the harness's fallback transmission path that ensures
graceful degradation rather than hard failure.

\textbf{Layer~4: Decision-to-Evidence Coupling (SO Decision Trace).}
The fourth transmission point is the accountability boundary: converting every
decision output into an ontologically typed, fully replayable evidence artifact whose operation sequence can be replayed by any auditor against the original EO graph snapshot.
This layer writes the complete execution trace---entity alignments with confidence
scores, scenario condition set $\mathcal{R}$ with EO provenance, Phase~2 sandbox
mutation log, Phase~3 skill invocation and result---to the session ontology as a
structured decision trace.
The evidence chain is the harness's final output: it is not a log file, but an
ontologically governed artifact that any auditor can replay against the original EO
graph snapshot to verify that the harness transmitted the ontology's authority
correctly at every step.

\subsubsection{Harness Quality Criterion}

A harness is well-engineered when it transmits the engine's power with minimal loss and
maximal precision.
For the ontology intelligence harness, this criterion translates directly:

\emph{Every operation the LLM performs should be traceable to an EO-authorized
constraint, and no operation should reach the LLM without first passing through the
appropriate harness layer.}

Violations of this criterion manifest in two characteristic failure modes.
\emph{Ontology bypass}: an operation proceeds without full EO grounding---an entity
enters the sandbox unresolved, a scenario condition is drawn from model-internal
knowledge rather than EO authority, or a skill is invoked without satisfying its
EO-linked preconditions.
The result is a fluent but unauthorized decision: the engine's power was not transmitted
through the harness, it leaked around it.
\emph{Ontology underuse}: the harness is simplified to the point where EO authority is
treated as optional context rather than mandatory coupling---the ontology is consulted
rather than enforced.
The result is a system that works in testing but degrades in production as edge cases
expose the gap between consultation and enforcement.

Both failure modes are detected by the tool-chain F1 metric: an agent that produces
correct answers without traversing all four harness layers receives F1 penalties
precisely because it has bypassed one or more transmission points.
The illusive accuracy phenomenon---high Acc with near-zero F1---is the empirical
signature of ontology bypass at scale: the model answers correctly, but the harness
has not transmitted the ontology's authority into the decision.

\subsection{The Extended Ontological Scope}

Traditional enterprise ontology systems govern only the EO layer: entity definitions,
relation schemas, constraint conditions, and access policies---the semantic authority
over what exists and what is permitted.
This coverage is necessary but insufficient for a system that must also govern what can
be done, how it is done, and in what execution context.
LOM extends the ontological scope to six coupled layers, each carrying a distinct
governance function yet all deriving their authority from the EO foundation.
These layers are organized by \emph{functional role}---what each layer stores, declares,
or enforces---rather than by abstraction level; they form a coupled ensemble, not a
strict hierarchy.

\textbf{Layer~1: Enterprise Ontology (EO).}
The permanent semantic authority of the enterprise.
It comprises five constraint families: \emph{EO.Standard-ID} canonical identifiers for
every entity and relation; \emph{EO business scenario conditions} encoding the
constraint predicates that restrict accessible entities and permissible operations;
\emph{EO.Authorization-Model} governing which users may invoke which skills;
\emph{EO.Logic-Constraint} specifying exact computation formulas for metrics; and
\emph{EO.Enumeration-System} defining legal value sets for categorical attributes.
The EO is read-only at inference time and is never modified by any pipeline phase.
It is the engine whose authority all downstream layers transmit.
Every entity, relation, and constraint entering the reasoning pipeline must be
EO-grounded before simulation proceeds.

\textbf{Layer~2: Bridge Layer Ontology.}
The semantic interface between raw enterprise inputs and the EO.
Rather than recording operations (which Section~A.4 covers), this layer stores the
\emph{artefacts} those operations produce: alignment mapping nodes pairing each surface
expression with its EO.Standard-ID canonical target; confidence-annotated triple records
carrying the alignment score $c \in [0,1]$ alongside the grounded $(s, p, o)$ triple;
and HITL interaction artefacts logging which candidate EO nodes were surfaced, which was
confirmed, and at what confidence threshold resolution was achieved.
All artefacts are typed bridge-layer ontology nodes persisted in the SO for full
provenance across every entity-grounding decision that precedes Phase~2 simulation.

In cross-domain enterprise deployments, alignment mapping nodes are naturally expressed
as OWL alignment axioms.
\texttt{owl:equivalentClass} is appropriate for strict semantic identity; when the
correspondence is directional or subject to drift, \texttt{rdfs:subClassOf} is the
safer primitive, preserving one-way inheritance without forcing bidirectional coupling.

\textbf{Layer~3: Skill Ontology.}
The capability registry of the harness.
Every registered capability---deterministic tool calls, computation engines, and
delegated frontier-LLM invocations---is declared as a typed node carrying formal
preconditions $\text{Pre}(s)$, postconditions, input/output type signatures, and
EO-linked authorization constraints.
The skill ontology is not a flat API registry: activation conditions derive from EO
graph traversal, invocation boundaries are enforced by EO.Authorization-Model, and
outputs are typed back into the EO namespace.

This layer also encodes \emph{implementation bindings}: concrete tool endpoints (ERP
query interfaces, document workflow triggers, approval-chain APIs, financial computation
services) are represented as implementation variants of their parent capability nodes,
with endpoint signatures, retry and fallback contracts, dependency relationships, and
versioning metadata all traceable to EO authority.
A single capability node (e.g., \texttt{shortest\_path}) may admit multiple bindings
across runtime environments; the harness ontology (Layer~6) governs which binding is
active for a given session.
LOM-action's 19-function graph API suite is the current instantiation of this layer.

\textbf{Layer~4: Context Graph.}
The session-scoped, mutable simulation substrate on which Phase~2 sandbox operations
and Phase~3 decision derivation execute.
Every node and edge in the context graph carries EO.Standard-ID provenance; every
mutation is a typed delta written under ontological governance.
The context graph is not a free-form scratchpad---it is an auditable extension of the
EO knowledge space, sandboxed to the current session via \texttt{graph\_id} and
constrained to the active scenario condition set~$\mathcal{R}$.
Within the six-layer scope, it serves as the \emph{live execution surface} of the EO:
where the EO encodes permanent organizational authority, the context graph encodes the
scenario-evolved state of that authority for the current business event.

\textbf{Layer~5: Memory Ontology.}
Conversational memory in LOM is an ontologically typed artefact, not raw text history.
The session ontology (SO) stores three artefact classes: \emph{graph deltas}---typed
triple mutations annotated with EO node references and turn indices;
\emph{turn logs}---per-turn execution traces recording active scenario conditions,
execution mode, and produced deltas; and \emph{decision traces}---the session's primary
governance deliverable spanning all turns.
All SO entries conform to the same ontological typing discipline as EO entities and
skill ontology nodes, enabling EO provenance to propagate transitively through every
memory artefact and making every recalled fact traceable to an authoritative source.

\textbf{Layer~6: Harness Ontology.}
The meta-governance layer that manages the lifecycle of the execution environment
itself: the sandbox session registry, the skill ontology registry, the HITL
clarification protocol, and the LOM-as-Judge evaluation contract.
The harness ontology encodes which execution modes are admissible for which event types,
which sandbox operations are authorized under which scenario conditions, and how the
four transmission layers (Section~A.1.3) are sequenced and enforced.
It is the layer that makes LOM a harness rather than a wrapper: it does not merely
route calls through the ontology---it governs the environment in which ontological
authority is coupled to every execution step.
Critically, the harness ontology is \emph{self-governing}: its schema is fixed at
design time and does not participate in the RAC evolutionary cycle
(Section~A.8), avoiding an infinite regress in which the governance layer would itself
require a meta-governance layer.

\medskip
Together, these six layers---EO, bridge layer ontology, skill ontology, context graph,
memory ontology, and harness ontology---constitute the full LOM ontological scope.
Their unification under a single ontology-management paradigm makes LOM's audit
guarantees \emph{transitive}: an authorization established at the entity level in
Layer~1 propagates consistently through skill activation (Layer~3), sandbox graph
mutation (Layer~4), memory persistence (Layer~5), and execution governance (Layer~6),
with every inference step traceable to the same ontological authority.
This transitivity is what distinguishes a system that produces plausible outputs from
one that produces auditable decisions.

\subsection{The Three Ontological Stores}

The EO layer is operationalized through three persistent stores with distinct access
semantics.
Together they form the horse that drives all LOM reasoning; no inference proceeds
without grounding in one of them.

\textbf{Enterprise ontology (EO) [global, read-only]} is the permanent semantic
authority.
It comprises five families of constraints and identifiers, each playing a specific role
in the CAR reasoning cycle: EO.Standard-ID provides canonical, unique codes for every
entity and relation; EO business scenario conditions encode the constraint predicates
that restrict which entities are accessible and which operations are permissible;
EO.Authorization-Model governs which users may invoke which skills;
EO.Logic-Constraint specifies exact computation formulas for metrics; and
EO.Enumeration-System defines the legal value sets for categorical attributes.
The EO is never modified at inference time. It is the horse.

\textbf{Personal ontology (PO) [per-user, read-only]} is a user-scoped semantic overlay
that carries alias mappings, preferred metric calibrations, and field suppression
preferences.
PO operates strictly within EO.Authorization-Model boundaries and never overrides any
EO constraint.

\textbf{Session ontology (SO) [per-session, read-write]} is the mutable simulation
substrate for the current conversation.
It stores three artifact types: graph deltas (typed triple additions and deletions
annotated with the EO nodes they reference and the execution mode that produced them);
turn logs (per-turn execution traces recording which scenario conditions were active,
which execution mode was used, and which graph deltas were produced); and decision
traces (the full audit trail across all turns, constituting the session's primary
governance deliverable).

\subsection{Semantic Parsing and Ontological Alignment}

\subsubsection{Natural Language to Semantics}

A foundational capability of LOM is the semantic parsing layer that converts both
user-facing natural language queries and system-level business scenario conditions into
ontology triples---the universal currency of the LOM reasoning engine.
This unification is architecturally significant: it means that user intent and
organizational scenario conditions are expressed in the same representation, enabling
them to be reasoned over jointly rather than reconciled at the prompt level.

\textbf{Event and input semantic parsing.}
A user's natural language input is parsed by the alignment function
$\text{Align}(q) \to \{(s_i, p_i, o_i, c_i)\}$, producing a set of typed ontology
triples $(s, p, o)$ where $s$ and~$o$ are EO.Standard-ID-anchored entities or values
and~$p$ is an EO-authorized predicate, each carrying a confidence score
$c_i \in [0, 1]$.

\textbf{Business scenario semantic parsing.}
Enterprise business scenarios undergo the same triple-grounding process.
A scenario such as ``only cost-center L2 nodes are traversable for this user's role''
is not injected as a natural language instruction into the prompt; it is parsed into
constraint triples and registered as active $\mathcal{R}$ predicates in the ontological
session state.
This is what makes Phase~2 sandbox simulation a computation over triples rather than a
pattern-matching exercise over scenario text.

\textbf{Precise subgraph localization through triple grounding.}
Because both the event payload and the active scenario condition set are expressed as ontology
triples anchored to EO.Standard-ID entities, subgraph localization reduces to a
well-defined graph operation: identify the subgraph $G_{\mathcal{R}}$ that satisfies
all constraint triples from $\mathcal{R}$ and contains all entity nodes referenced by
the event payload triples.
This triple-to-subgraph mapping is exact and deterministic.

\subsubsection{The LOM Interface Contract}

The semantic parsing layer defines LOM's external interface contract.
LOM is a bidirectional semantic transducer that accepts inputs and produces outputs in
two representation registers---natural language and formal ontology (OWL/triple
notation)---independently on each side:
$$\text{NL} \;\Big|\; \text{OWL} \;\xrightarrow{\quad\text{LOM}\quad}\; \text{NL} \;\Big|\; \text{OWL}$$

This yields four canonical interaction modes, summarized in Table~\ref{tab:io-mapping}:

\begin{table*}[h]
\centering
\caption{Mapping of Input/Output Modalities and Use Cases}
\label{tab:io-mapping}
\begin{tabularx}{\textwidth}{llX}
\toprule
\textbf{Input} & \textbf{Output} & \textbf{Typical Use Case} \\
\midrule
Natural Language & Natural Language & Business event or user request $\to$ human-readable decision with evidence narrative \\
\addlinespace
Natural Language & OWL / Triples    & Business event or user request $\to$ machine-readable ontology delta for downstream systems \\
\addlinespace
OWL / Triples    & Natural Language & Formal scenario or graph state $\to$ explanation, audit report, clarification dialogue \\
\addlinespace
OWL / Triples    & OWL / Triples    & System-to-system ontology enrichment, automated scenario verification, RAC-driven EO update \\
\bottomrule
\end{tabularx}
\end{table*}

\subsection{The CAR--RAC Closed Loop: Forward Execution and Evolutionary Feedback}

LOM operates through a self-reinforcing closed loop composed of two coupled cycles.
CAR (Construct $\to$ Align $\to$ Reason) is the \emph{forward pass}: it transforms
raw enterprise data and incoming business events into auditable, ontologically grounded
decisions.
RAC (Reason $\to$ Align $\to$ Construct) is the \emph{feedback pass}: it is the exact
reverse of CAR, scanning every forward-pass output for organizational knowledge not yet
formally represented in the EO and routing validated candidates back into the ontology
as permanent entries.
Together they form a single governed flywheel---each forward decision strengthens the
ontological substrate that future forward decisions depend on.

\subsubsection{CAR: The Forward Execution Pipeline}

CAR transforms raw enterprise inputs into auditable decisions through three
EO-constrained steps.

\textbf{Construct} is the autonomous ontology construction step.
Rather than requiring a hand-crafted schema, LOM ingests raw enterprise data and
autonomously identifies entity types, relation types, constraint predicates, and
canonical identifiers.
EO.Enumeration-System constrains the legal value sets for categorical attributes
discovered during construction; EO business scenario conditions govern which entity
classes and relation types are admissible for the current organizational scope.
The output is a typed ontological graph whose nodes and edges carry EO.Standard-ID
provenance---the structural substrate on which all downstream alignment and reasoning
operate.

\textbf{Align} performs two coupled operations: dynamic ontology update and
multimodal ontology--language alignment.
On the update side, newly constructed ontological elements are reconciled with the
existing EO---resolving synonym merges, deduplicating entities, and propagating
constraint inheritance---so the ontology remains consistent as new data arrives.
On the alignment side, a graph-aware encoder bridges the structural ontology
representation with the natural language surface of user queries and business events,
mapping free-form language onto EO.Standard-ID-anchored entities and EO-authorized
predicates.
The ordering within Align is architecturally enforced: EO constraints are resolved
first, PO preferences applied second---never reversed.
EO.Authorization-Model is the non-bypassable arbiter for all entity access decisions.

\textbf{Reason} executes deterministic inference over the constructed and aligned
ontological topology, operating on node attributes, relation types, and the constraint
predicates established in the preceding two steps.
Every inference step produces a confidence score written to the SO alongside its
result.
Steps falling below the configurable confidence threshold trigger the compliance gate:
the system either requests clarification, escalates to a registered top-tier LLM
skill, or issues a structured refusal---never silently continuing with a low-confidence
inference.
High-confidence results are written to the SO decision trace with full EO node
provenance, forming the auditable output of the forward pass.

\subsubsection{RAC: The Evolutionary Feedback Pass}

RAC runs as the reverse cycle after every CAR execution, closing the loop between
deployment experience and ontological knowledge.

Specifically, RAC scans the SO decision trace for entities, relations, metric names,
and scenario patterns whose confidence scores exceed the acceptance threshold but which
are not yet formally represented in the current EO.
Each such element is captured as a typed candidate node annotated with its
provenance---the turn that produced it, the execution mode used, and the SO delta in
which it appears---and routed to a governance queue for enterprise knowledge manager
review.

Validated candidates are promoted to permanent EO entries via the Construct step,
expanding the canonical identifier namespace and tightening the constraint space.
The updated EO then feeds a new Align pass, making the ontology--language alignment
function more precise for future sessions.
This creates the flywheel property: high-quality forward decisions produce
high-confidence SO artefacts; high-confidence artefacts surface well-evidenced EO
candidates; validated candidates sharpen the ontology that governs future forward
decisions.

Three properties of the RAC cycle are architecturally significant.
First, \emph{provenance preservation}: every candidate node carries a complete
lineage---which session, which turn, which execution mode, which SO delta---so that
any promoted EO entry can be retrospectively audited against the interaction that
generated it.
Second, \emph{governed promotion}: candidates never enter the EO autonomously;
the human governance review queue is a mandatory gate, ensuring the ontology grows
from deployment interactions through a versioned, auditable process rather than
through unchecked model inference.
Third, \emph{graceful expansion}: because the harness ontology (Layer~6) is
self-governing and excluded from the RAC cycle, the evolutionary flywheel cannot
corrupt its own governance mechanism---the loop is closed at the knowledge layer,
not at the control layer.

The ontology is the horse; RAC ensures the horse grows stronger with every ride
without ever changing the rules by which it is guided.







\subsection{Ontological Gating and Authorization Control}

LOM enforces a two-stage safety envelope around every reasoning pipeline execution.
The first stage---ontological injection and confidence gating---operates at the input
boundary, ensuring that no entity enters the pipeline without EO grounding.
The second stage---scenario and authorization hard stops---operates throughout all CAR
steps, halting execution whenever an EO constraint is violated during simulation or
decision derivation.
Together, they constitute the harness's non-bypassable correctness guarantee.

\subsubsection{Ontological Injection and Confidence Gating}

Before CAR begins, every incoming input undergoes \emph{ontological injection}: the
system resolves entities to EO.Standard-ID canonical codes---building on entity
recognition foundations established by neural sequence labeling
architectures~\cite{zhu2019flexner}---validates values against EO.Enumeration-System,
injects applicable EO.Logic-Constraint into the semantic representation, and surfaces
PO alias preferences.

Each entity resolution produces a confidence score $c \in [0, 1]$ from the alignment
function $\text{Align}(q) \to (v, c)$.
Three thresholds govern what happens next.
If $c \geq \theta_{\text{accept}}$, the entity is grounded and reasoning proceeds
immediately.
If $\theta_{\text{clarify}} \leq c < \theta_{\text{accept}}$, a targeted
clarification question is generated and the pipeline is held until the user confirms
or corrects the candidate EO node.
If $c < \theta_{\text{clarify}}$, the expression is flagged as an ontology gap
candidate: it is written to the SO as a typed candidate node for the RAC evolutionary
cycle (Section~A.8), and execution is suspended until the gap is resolved.

This confidence-gated alignment ensures that no entity enters the reasoning pipeline
at below-threshold precision---and that every grounding decision is itself a typed,
auditable SO artefact with full EO provenance.

\subsubsection{Scenario and Authorization Hard Stops}

Three hard stops apply throughout all CAR steps, each triggering an immediate,
structured halt rather than a silent degradation.

\textbf{Stop~1: Entity non-existence} (\texttt{EO.Standard-ID} not found).
The incoming event references an entity with no corresponding EO node.
Execution halts; a targeted clarification citing the nearest ontological neighbours
is returned to the requester.

\textbf{Stop~2: Permission violation} (\texttt{EO.Authorization-Model} check failed).
The requested skill invocation or data access falls outside the user's authorized scope.
Execution halts; a structured refusal citing the specific permission boundary is
returned---no partial result is written to the sandbox.

\textbf{Stop~3: Calibration conflict} (\texttt{EO.Logic-Constraint} mismatch).
The requested computation would apply a metric calibration inconsistent with EO
standards.
Execution halts; the conflict is surfaced for explicit resolution before any
numerical inference proceeds.

In all three cases, the halt itself writes a decision trace entry to the SO: failed
queries are fully auditable, not silently discarded.
This means the audit trail is complete regardless of whether execution succeeds or
stops---a property essential for regulated enterprise environments where the absence
of a decision carries as much compliance significance as the decision itself.

\subsection{From Reasoning to Action: LOM-action as the Operative Extension of LOM}

\subsubsection{The Capability Progression: Reason $\to$ Act}

LOM's foundational contribution is ontologically grounded \emph{reasoning}: the CAR
pipeline transforms raw enterprise data into auditable decisions whose every inference
step is traceable to an EO-authorized constraint.
LOM-action extends this foundation along a single, precise axis---from
\emph{reasoning about} the enterprise ontology to \emph{acting upon} it.

The distinction is architectural, not merely functional.
In pure reasoning mode, LOM produces conclusions: it reads the ontological graph,
applies EO-encoded constraints, and returns a grounded answer.
In action mode, LOM-action produces \emph{mutations}: it receives a business event,
simulates its scenario conditions as deterministic graph operations in an isolated
sandbox, and derives a decision exclusively from the evolved graph state---a decision
that is itself a typed, replayable operation on the ontological substrate.
The reasoning capability is not replaced; it is extended.
LOM-action is LOM with an operative harness attached---a harness that converts
ontological authority into directed, auditable action rather than grounded text.

This progression can be stated compactly as:
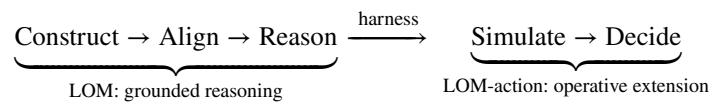
\begin{align*}
\underbrace{\text{Construct} \to \text{Align} \to \text{Reason}}_{\text{LOM: grounded reasoning}}
\;\xrightarrow{\text{harness}}\;
\underbrace{\text{Simulate} \to \text{Decide}}_{\text{LOM-action: operative extension}}
\end{align*}

The CAR cycle produces the ontologically grounded knowledge state that the harness
requires as its input.
The harness---realized as the event $\to$ simulation $\to$ decision pipeline described
in Section~3---converts that state into action: it instantiates an isolated sandbox,
applies EO-authorized scenario conditions as deterministic graph mutations, and derives
every downstream decision exclusively from the resulting simulation-valid graph
$G_{\text{sim}}$.
Neither half is sufficient alone.
CAR without the harness produces grounded reasoning with no operative output.
The harness without CAR produces sandbox mutations with no ontological authority.
LOM-action is their coupling: the point at which enterprise knowledge becomes
enterprise action.

\subsubsection{The Onto OS Vision: An Ontological Operating System for the Enterprise}

The six-layer ontological scope (Section~A.2), the CAR--RAC closed loop
(Section~A.6), and the LOM-action operative harness are not independent engineering
contributions.
They are three components of a single long-horizon architectural trajectory whose
terminal destination is an \emph{Ontological Operating System}---Onto~OS.

An operating system governs a computational environment: it manages resources,
enforces access control, schedules execution, and provides a stable interface through
which applications interact with hardware without requiring direct hardware access.
Onto~OS applies this paradigm to enterprise knowledge and decision infrastructure.
Its governed resource is the enterprise ontology; its access control layer is
EO.Authorization-Model; its execution scheduler is the harness ontology; its stable
interface is the skill ontology registry through which every capability---tool calls,
LLM invocations, computation engines---interacts with organizational authority without
bypassing it.

Concretely, Onto~OS is the system in which:

\begin{itemize}
\item \emph{Every business event} is an operating system call---a structured payload
that activates EO-encoded scenario conditions, enters the simulation pipeline, and
produces an auditable decision trace as its return value.

\item \emph{Every capability invocation} is an OS-mediated system call---a skill
ontology node whose preconditions are EO-grounded, whose execution is sandbox-isolated,
and whose output is typed back into the EO namespace before any downstream system
receives it.

\item \emph{Every reasoning step} is a governed process---scheduled by the harness
ontology, constrained by EO.Logic-Constraint, and logged to the SO with full
provenance, so that no inference executes outside the ontological authority of the
enterprise.

\item \emph{The enterprise ontology itself} evolves continuously through the RAC
flywheel---a versioned, governance-gated process analogous to a kernel update: each
promotion expands the capability surface of the OS while preserving the integrity
guarantees of all prior sessions.
\end{itemize}

Under Onto~OS, the enterprise is not an information source to be queried---it is a
continuously operating event environment whose data streams and business events arrive
as structured system calls into the ontological pipeline.
Each incoming event activates EO-encoded scenario conditions, enters the simulation
sandbox, and exits as an auditable decision trace---not a response to a human prompt,
but a governed operative output of the ontological runtime.
The frontier LLM is neither the engine nor the authority---it is one registered
capability among many, invoked under EO authorization when no smaller LOM variant can
cover the required computation, evaluated by LOM-as-Judge before its output is typed
back into the ontological substrate, and never permitted to influence any downstream
decision outside the simulation boundary.
Model scale becomes a capability parameter, not an architectural dependency.

LOM-action is the first operative instantiation of this vision.
Its 19-function graph API suite, sandbox simulation engine, dual-mode execution
architecture, and decision trace infrastructure are the kernel primitives of Onto~OS
in their current, graph-domain form.
The path from this prototype to a full enterprise Onto~OS runs through the four future
work axes identified in Section~5: SKILLS-standard integration to replace
natural-language scenario descriptions with formal ontological declarations; RAC
governance hardening to provide versioned audit-chain validity scopes; cross-scale and
cross-domain validation to demonstrate transfer beyond synthetic graph ontologies; and
latency characterization to establish event-throughput guarantees under production
load.

The architectural claim is precise: \emph{trustworthy enterprise AI is not a property
of model scale---it is a property of ontological governance}.
Onto~OS is the environment in which that governance is fully instantiated.
LOM-action is its first working component.

\end{document}